\documentclass[10pt]{article} 

\usepackage[accepted]{tmlr}

\usepackage{hyperref}
\usepackage{url}
\usepackage{amsthm,amssymb}
\usepackage{dsfont}
\usepackage[most]{tcolorbox}
\usepackage{color}
\usepackage{multicol}
\usepackage{enumitem}
\usepackage{thm-restate}
\usepackage{cases}

\usepackage{lipsum}

\usepackage[most]{tcolorbox}
\newtcolorbox{nbox}[1][]{
  enhanced,
  fonttitle=\scshape,
  #1
}

\title{Policy Optimization via Adv2: \\ Adversarial Learning on Advantage Functions}

\author{\name Matthieu Jonckheere \email matthieu.jonckheere@laas.fr \\
     \addr LAAS, Université de Toulouse, CNRS, Toulouse, France
     \AND
     \name Chiara Mignacco \email chiara.mignacco@universite-paris-saclay.fr \\
     \addr Université Paris-Saclay, CNRS, Inria, Laboratoire de mathématiques d'Orsay, 91405, Orsay, France
     \AND
     \name Gilles Stoltz \email gilles.stoltz@universite-paris-saclay.fr \\
     \addr Université Paris-Saclay, CNRS, Inria, Laboratoire de mathématiques d'Orsay, 91405, Orsay, France}

\def\month{05}  
\def\year{2025} 
\def\openreview{\url{https://openreview.net/forum?id=Oyueig10Ed}} 

\renewcommand{\geq}{\geqslant}
\renewcommand{\leq}{\leqslant}
\renewcommand{\epsilon}{\varepsilon}

\newcommand{\E}{\mathbb{E}}
\renewcommand{\P}{\mathbb{P}}
\newcommand{\R}{\mathbb{R}}

\newcommand{\cS}{\mathcal{S}}
\newcommand{\cA}{\mathcal{A}}
\newcommand{\cP}{\mathcal{P}}
\newcommand{\cX}{\mathcal{X}}
\newcommand{\cC}{\mathcal{C}}
\newcommand{\cR}{\mathcal{R}}

\newcommand{\cT}{\mathcal{T}}

\newcommand{\bDelta}{\boldsymbol{\Delta}}
\newcommand{\bdelta}{\boldsymbol{\delta}}
\newcommand{\bpi}{\boldsymbol{\pi}}
\newcommand{\bcR}{\boldsymbol{\cR}}
\newcommand{\bcT}{\boldsymbol{\cT}}
\newcommand{\br}{\boldsymbol{r}}
\newcommand{\bq}{\boldsymbol{q}}
\newcommand{\bp}{\boldsymbol{p}}

\newcommand{\proj}{\mathop{\mathrm{proj}}}

\newcommand{\eqdef}{\stackrel{\mbox{\scriptsize \rm def}}{=}}
\newcommand{\wh}{\widehat}

\newtheorem{theorem}{Theorem}
\newtheorem{example}{Example}
\newtheorem{definition}{Definition}
\newtheorem{lemma}{Lemma}
\newtheorem{remark}{Remark}
\newtheorem{corollary}{Corollary}
\newtheorem{assumption}{Assumption}
\newtheorem{oracle}{Oracle}

\newcommand{\contrib}[1]{\paragraph{Contributions of this section.} \emph{#1}}

\begin{document}

\maketitle

\begin{abstract}
We revisit the reduction of learning in adversarial
Markov decision processes [MDPs] to adversarial learning based on $Q$--values;
this reduction has been considered in a number of recent articles as one building block to perform policy
optimization. Namely, we first consider and extend this reduction in an ideal setting where
an oracle provides value functions:
it may involve any adversarial learning strategy (not just exponential weights)
and it may be based indifferently on $Q$--values or on advantage functions.
We then present two extensions: on the one hand, convergence of the last iterate
for a vast class of adversarial learning strategies (again, not just exponential weights),
satisfying a property called monotonicity of weights;
on the other hand, stronger regret criteria for learning in MDPs, inherited from the stronger regret
criteria of adversarial learning called strongly adaptive regret and tracking regret.
Third, we demonstrate how adversarial learning, also referred to as aggregation of experts,
relates to aggregation (orchestration) of expert policies:
we obtain stronger forms of performance guarantees in this setting than existing ones,
via yet another, simple reduction.
Finally, we discuss the impact of the reduction of learning in adversarial
MDPs to adversarial learning in the practical scenarios
where transition kernels are unknown and value functions must be learned.
In particular, we review the literature and note that many strategies for policy optimization
feature a policy-improvement step based on exponential weights with estimated
$Q$--values. Our main message is that this step may be replaced by the application
of any adversarial learning strategy on estimated $Q$--values
or on estimated advantage functions. We leave the empirical evaluation
of these twists for future research.
\end{abstract}

\section{Introduction}

In this article, we revisit
a specific approach in policy optimization
for adversarial Markov decision processes [MDPs] in the episodic setting,
namely, the closed-form design of policies selected over time (which change incrementally)
based on estimated value functions. In virtually all previous work,
these policies are computed via the same adversarial-learning strategy,
referred to under possibly different names: exponential weights, weighted majority,
Boltzmann reweighting, or online mirror descent, to name a few.
It turns out that different adversarial-learning strategies may be
used, which may have important consequences in practice.

Put differently,
this article aims to formally establish the mathematical consistency between the study of adversarial MDPs and (plain) adversarial learning,
and to effectively bridge these two important areas of learning theory.

\subsection{Brief literature review}
\label{sec:lit-review}

Before reviewing in detail our contributions, we first provide
a concise overview of the related literature and justify
some claims contained in the previous paragraph.

\paragraph{Adversarial MDPs / Reduction to adversarial learning.}
The setting of adversarial MDPs was introduced
by~\cite{ED09} and~\cite{Yu09}. As
in the standard episodic setup, the transition kernels dictating the evolution
of the states are unknown and constant over time. However, the reward functions
vary over time and may be chosen by some adversary; they are possibly revealed
at the end of an episode. Both references were also the first ones to introduce a reduction
of the control of adversarial MDPs to standard adversarial learning (a setting also called
expert prediction; see \citealp{CBL06} for an overview thereof).
In this article, we will be interested in closed-form policy optimization,
and not in approaches relying on so-called occupancy measures (introduced by \citealp{ZN13}), which
solve a complex convex optimization problem at each episode, without resulting in closed-form expressions for the output policies (see, e.g., \citealp{RM19}).

\paragraph{Policy optimization.}
Policy optimization refers to designing policies to be used at each episode,
often obtained by sequential incremental updates, and may be opposed to
value-based learning in MDPs, which focuses on estimating and improving value functions rather than directly constructing policies.
Several approaches were considered in policy optimization, for instance, (natural) policy gradient
(\citealp{Sutton00}, \citealp{KL02}), and variants like Trust Region Policy Optimization or Proximal Policy Optimization
(TRPO and PPO, respectively; see \citealp{Schu15}, \citealp{Schu17}).
We will rather be interested in the closed-form policy design
relying on estimates of $Q$--value functions.
This vein of research includes the works by \citet{Shani20},
\citet{Cai20}, \citet{He22}, \citet{Zhao23}, \citet{Tiapkin25} to name a few representative contributions (see also \citealp{Politex19}). The settings differ in these articles
depending, among others, on the feedback on the reward functions (full monitoring or bandit feedback) and on the
structural assumptions, or lack thereof, on the transition kernels.

However, all cited references have one thing in common: they rely on the same adversarial-learning strategy
to process the estimated $Q$--value functions
(except \citealp{Tiapkin25}, which builds on the present work).

\paragraph{A single adversarial-learning strategy, based on exponential weights.}
This same adversarial-learning strategy is known under different names
and relies on exponential weights;
for instance, \citet[Section~5.3]{AKLM21} refers to it as multiplicative weights updates,
\citet{Politex19}, as the Boltzmann policy\footnote{\citet{Politex19} even states that
``the choice of the Boltzmann policy is not arbitrary'', but one of the points of the present article
is to actually show the contrary: many other choices of adversarial-learning strategies are suitable.},
\citet{Shani20} and \citet{Zhao23}, as online mirror descent (with a Kullback-Leibler
regularization).
\citet{Cai20} and \citet{He22} do not write any explicit strategy name
for the corresponding step of their policy-optimization approach,
but refer to the same closed-form update considered by
earlier references; they obtain it
by resorting to some follow-the-regularized-leader approach
with an entropic regularization (known to
lead to exponential weights, see \citealp{FSSW97}, \citealp{KW99},
\citealp{Aud09}).

Interestingly, this strategy based on exponential weights aligns with the concept of natural policy gradient for non-adversarial MDPs when the policy parametrization is softmax: both approaches involve the same update rule on the weights
(this explicit update rule was, for instance, derived in \citealp[Section~5.3]{AKLM21}, see also \citealp{Kak01}).
This specific case, as the intersection of two optimisation paradigms, leads to remarkable theoretical guarantees in non-adversarial MDPs;
see, in particular, the recent work by \cite{MM24} and references therein.

Two exceptions to the use of the exponential-weight strategy are provided
by~\cite{ED09} and~\cite{Yu09}, which resort to a strategy
called follow-the-perturbed-leader (\citealp{KV05});
but their setting and objectives are somewhat different from those considered in this article and the previous references.

\paragraph{Previous reductions of learning in MDPs to adversarial learning.}
We provide a specific analysis of the strategy based on exponential weights in Section~\ref{sec:NPG},
obtaining improved regret bounds compared to the analyses provided
in the mentioned references. These analyses
range from a few-line-long proof performing a direct reduction to adversarial learning in \citet{Shani20} (a proof that we copy in Section~\ref{sec:advL-adv}
but that can be improved in the specific case of exponential weights),
to longer proofs (possibly several pages, see, e.g., \citealp[Appendix~A.1]{Zhao23}).
The typical proofs are between these two extremes. In particular,
to the best of our knowledge, no other proof than the one by \citet{Shani20} clearly identifies a reduction, and
all other proofs rather mimic and adapt\footnote{Typical proofs usually consider the analysis of
exponential weights based on telescoping Kullback-Leibler terms (as in \citealp{FS99}), but we note that shorter analyses exist, e.g.,
based on Hoeffding's lemma (see \citealp[Section~2.2]{CBL06}).} the analysis of exponential weights in
adversarial learning, as in \citet[Section~5.3]{AKLM21} or \citet{Cai20}.
We note that the cited references actually run the exponential-weight strategy
on estimated $Q$--values: more details are provided in Section~\ref{sec:EstAdv}.

In a nutshell, among all cited references, \citet{Shani20} already clearly identified
how to reduce learning in MDPs to adversarial learning, but only leveraged this fact
for one specific adversarial learning strategy.

\subsection{Contributions and outline of this article}

In Section~\ref{sec:setting-main},
we formally define the setting of episodic adversarial MDPs
and state our objective: the minimization of a cumulative regret, defined as the sum of the differences
between the value functions of the best stationary policy and of the output policies.

Section~\ref{sec:oracle} recalls the reduction of learning in MDPs to adversarial learning
as clearly stated in the course of a proof by~\citet{Shani20}.
We state the reduction in the ideal setting where an oracle provides, at the end of each episode, the value functions corresponding to the policy played---a restriction that we discuss
and mitigate later in Section~\ref{sec:EstAdv}.
We essentially replicate the proof by~\citet{Shani20},
based on the performance difference lemma, and actually generalize it by
considering a broad family of possible adversarial learning strategies, not
just exponential weights with a constant learning rate.
For instance, the \texttt{ML-Prod} and \texttt{ML-Poly} strategies (\citealp{Gai14}, \citealp{CRAN}) are suitable adversarial-learning
strategies that exhibit in general much better empirical performance
than exponential weights.
Another observation is that the theoretical guarantees hold when
adversarial-learning strategies are fed with advantage functions instead of $Q$--functions, which constitutes a second possible source of improved empirical performance.

We then discuss three extensions (convergence of the last iterate, stronger
forms of regret, aggregation of policies) and present two twists: a special-case analysis
for exponential weights with improved bounds, and how to use the general theory developed
in practical scenarios where advantage functions must be estimated (in particular due to the transition
kernels being unknown).

\paragraph{Extension 1: convergence of the last iterate.}
In the case where reward functions are constant over time,
Section~\ref{sec:monot} focuses on a regret called simple regret, which measures
the difference in performance between the best stationary policy and the last policy selected.
\citet[Section~5.3]{AKLM21} controlled this quantity for exponential weights
with a constant learning rate (in the discounted setting). We show how to extend their
argument to a large class of adversarial strategies satisfying a natural property
that we call ``monotonicity of weights''.

\paragraph{Extension 2: Stronger forms of regret.}
Section~\ref{sec:stronger-regret} shows that the general reduction studied in Section~\ref{sec:oracle} also works
for a stronger notion of regret called strongly adaptive regret and consisting
of studying the sums of differences in value functions over sub-intervals of time.
As a consequence, the so-called tracking regret may also be controlled: therein, the comparison
is made not to the best stationary policy, but to the best sequence of policies with few shifts.
To the best of our knowledge, the control of such improved forms of regret for MDPs
is an original contribution.

\paragraph{The special case of exponential weights.}
Section~\ref{sec:NPG} leverages elements from Extensions~1 and~2 to show that
when the adversarial learning strategy consists in using exponential weights with a constant learning
rate, the (cumulative) regret may be bounded by the number of shifts in the reward sequence.
This provides yet another generalization of the results of \citet[Section~5.3]{AKLM21}.
In addition, the proof technique by \citet[Section~5.3]{AKLM21} seemed highly specific to
the discounted setting: we provide instead a treatment for the episodic setting.

\paragraph{Extension 3: Aggregation (orchestration) of expert policies.}
Adversarial learning is sometimes called prediction with experts (see \citealp{CBL06}).
Section~\ref{sec:aggreg} considers the case where policies selected over time are no longer learned in a
direct tabular setting, but are obtained by (state-by-state and stage-by-stage) convex
combinations of some expert policies. The aim is to mimic the performance of the overall best
such convex combination. This methodology, which we call aggregation (or orchestration) of expert
policies, is also referred to as learning from multiple oracles (which may be understood as a specific paradigm in the vast imitation-learning literature);
see, for instance, \citealp{CKA20} and \citealp{Liu23}. We show that to address this problem,
it suffices to consider expert policies as actions in a lifted MDP and
apply all results described earlier in this article. We obtain stronger
performance guarantees than in the cited references.

\paragraph{Empirical impacts as future research directions.}
Section~\ref{sec:EstAdv} puts into perspective the design of policies
studied in this article: in practice, advantage functions are unknown but may be estimated,
so that the strategies studied earlier in this article should be run on these estimates.
We review the literature of policy optimization to explain how and why the strategies
proposed in the literature may be modified: their policy-improvement step, stated with
exponential weights, may in fact rely on many other
adversarial-learning strategies. This modification has no impact on the
theoretical guarantees but could be impactful on the practical performance.
We however leave the assessment of that potential practical impact for future
research.

\section{Setting and aims}
\label{sec:setting-main}

\paragraph{Notation.}
We denote by $\cP(\cX)$ the set of probability distributions over some set $\cX$,
either finite or given by an interval of $\R$ in the sequel.
For an integer $n \geq 1$, let $[n] = \{1,\ldots,n\}$ denote the set
of the first $n$ integers.

\paragraph{Setting.}
We consider an $H$--episodic and (obliviously) adversarial Markov decision process [MDP]
with finite state and action spaces $\cS$ and $\cA$, of respective cardinalities $S$ and $A$:
each episode~$t \geq 1$ is of length $H \geq 1$ and is governed by transition kernels
$\bcT = (\cT_{h})_{h \in [H-1]}$, where $\cT_{h} : \cS \times \cA \to \cP(\cS)$,
and by reward functions $\bcR_t = (\cR_{t,h})_{h \in [H]}$, where
$\cR_{t,h} : \cS \times \cA \to \cP\bigl([0,1]\bigr)$.
The transition kernels are constant across episodes, while the reward functions
$\bcR_t$ vary between episodes; they may actually be picked by an adversary
in an oblivious manner, i.e., the entire sequence $(\bcR_t)_{t \geq 1}$
is determined by the adversary before the first episode takes place.

We denote by $r_{t,h} : \cS \times \cA \to [0,1]$
the mean-payoff function associated with $\cR_{t,h}$, i.e.,
$r_{t,h}(s,a)$ is the expectation of the distribution $\cR_{t,h}(s,a)$,
for each $s \in \cS$ and $a \in \cA$.

A (stationary, or one-shot) policy $\bpi = (\pi_h)_{h \in [H]}$ is a sequence of mappings $\pi_h : \cS \to \cP(\cA)$;
we denote by $\pi_h(\,\cdot\, | s)$ the probability distribution over actions that the policy uses in stage $h$ and state $s$.
The learner should determine a policy $\bpi_t$ at the beginning of each episode $t \geq 1$,
based on the information gained from rounds $\tau \leq t-1$; that information includes at least the states observed and actions played
therein, as well as the rewards obtained. In some scenarios, additional observations may be performed,
which we will explicitly detail; for instance, the learning system
may observe, among other things, the mean-payoff functions $\br_{\tau} = (r_{\tau,h})_{h \in [H]}$ at the end of episode~$\tau$.

At the beginning of each episode $t \geq 1$,
the same initial state $s_{t,1} = s_1$ is set.
Then, at each stage $h \in [H-1]$, the learning system draws an action
$a_{t,h} \sim \pi_{t,h}(\,\cdot\, | s_{t,h})$, after which
it obtains and observes a stochastic reward drawn independently from $\cR_{t,h}(s_{t,h},a_{t,h})$,
while the environment
moves to a new state drawn as $s_{t,h+1} \sim \cT_h(\,\cdot\,|s_{t,h},a_{t,h})$.
In the final stage, only an action $a_{t,H} \sim \pi_{t,H}(\,\cdot\, | s_{t,H})$
is drawn, and a reward drawn independently from $\cR_{t,H}(s_{t,H},a_{t,H})$ is obtained and observed.
We do not introduce pieces of notation for the rewards actually obtained as all arguments
in this article will be based on value functions, which, by the tower rule,
only depend on the mean-payoff functions.

More precisely, by the tower rule, the value function $V_h^{\bpi,\bcR_t}$ of a given stationary policy $\bpi = (\pi_{j})_{j \in [H]}$
at episode $t \geq 1$ and started at stage $h \in [H]$ equals, for all $s \in \cS$,
\begin{equation}
\label{eq:def-V}
V_h^{\bpi,\bcR_t}(s) = \E^{\bpi,\bcT} \! \left[ \sum_{j = h}^H r_{t,j}(s_j,a_j) \,\bigg|\, s_h = s \right],
\end{equation}
where the piece of notation $\E^{\bpi,\bcT}$ indicates that actions $a_h$
and states $s_h$ in the expectation are governed by the policy $\bpi$ and the transition kernels $\bcT$,
as described above.

\subsection{First aim: direct tabular learning}
\label{sec:direct}

We evaluate the policies $\bpi_t$ picked over time in terms of their value functions
and are interested in mimicking the performance of the best stationary policy in
hindsight. More precisely, the learning system aims to control
\begin{equation}
\label{eq:def-regret-V}
\forall T \geq 1, \qquad R_T = \max_{\bpi} \sum_{t=1}^T \Bigl( V_1^{\bpi,\bcR_t}(s_1) - V_1^{\bpi_t,\bcR_t}(s_1) \Bigr)\,,
\end{equation}
where the maximum is over all stationary policies~$\bpi$.
We write ``$\forall T \geq 1$'' to indicate that either the time horizon $T$ is unknown or
the regret should be controlled for all time horizons.
The regret $R_T$ involves a sum essentially because the reward functions $\bcR_t$ evolve over time
in a possibly adversarial way; when they are constant over time,
then convergence of the last iterate (i.e., of the $T$--th term in the sum above) may be achieved,
see Section~\ref{sec:monot}.

The aim described above is called direct tabular learning as policies $\bpi_t$
are picked by determining, for each stage $h$ and state $s$, the
entire probability distribution $\pi_{t,h}(\,\cdot\, | s)$.
The terminology is borrowed from \citealp[Section~3]{AKLM21}.

The setting above is summarized in Box~A.
\begin{figure}[t]
\begin{nbox}[title={Box~A: Policy optimization, for direct tabular learning}]
\ \\[-.15cm]
\textbf{MDP parameters:} state space $\cS$, action space $\cA$, initial state $s_1 \in \cS$,
transition kernels $\bcT$ \medskip

\textbf{Initialization:} The environment picks a sequence $(\bcR_t)_{t \geq 1}$ of reward functions \medskip

\textbf{For episodes} $t = 1, 2, \ldots$\textbf{:}
\begin{enumerate}[leftmargin=1cm,itemsep=1pt,topsep=3pt]
\item The initial state is set to $s_{t,1} = s_1$
\item \textbf{For stages} $h = 1, \ldots, H$\textbf{:}
\begin{enumerate}[label=(\alph*),topsep=0pt,itemsep=1pt,leftmargin=0.5cm]
\item The learner picks a policy $\pi_{t,h} : \cS \to \cP(\cA)$
\item and draws an action $a_{t,h} \sim \pi_{t,h}(\,\cdot\,|s_{t,h})$
\end{enumerate}
\item The learner receives and observes a reward drawn independently from $\cR_{t,h}(s_{t,h},a_{t,h})$,
with conditional expectation $r_{t,h}(s_{t,h},a_{t,h})$
\item If $h \leq H-1$, the next state $s_{t,h+1} \sim \cT_{h}(\,\cdot\,|s_{t,h},a_{t,h})$ is drawn
\end{enumerate}
\medskip

\textbf{Goal:} Minimize the regret $\displaystyle{R_T = \max_{\bpi} \sum_{t=1}^T \Bigl( V_1^{\bpi,\bcR_t}(s_1) - V_1^{\bpi_t,\bcR_t}(s_1) \Bigr)}$
\end{nbox}
\end{figure}

\paragraph{Alternative aim in Section~\ref{sec:aggreg}.}
The aim described in Box~A may be difficult to complete when the number $A$ of actions is large.
In addition, the learning system may sometimes have some prior
information given by a finite set of expert policies among which some policies could perform well
(the subsets of these good-performing policies could possibly depend on the state).
We therefore introduce an alternative aim in Section~\ref{sec:aggreg} called aggregation (or orchestration)
of expert policies, but actually show
that resolving this objective is equivalent in some sense to the aim described in Box~A.

\subsection{Additional notation}

For later use, we define $Q$--values and advantage functions, and use the same notation as in~\eqref{eq:def-V} to that end.
For any pair of stationary policy $\bpi$ and reward functions $\bcR$,
we define its $Q$--value function at episode $t \in [T]$, and started from stage $h \in [H]$, as
\[
Q_{h}^{\bpi,\bcR_t} : (s, a) \in \cS \times \cA \longmapsto
\E^{\bpi,\bcT} \! \left[\sum_{j = h}^H r_{t,j}(s_{j}, a_{j}) \biggm| s_{h} = s, \ a_{h} = a\right],
\]
and its advantage function as
\begin{equation}
\label{eq:def-A}
A_{h}^{\bpi,\bcR_t} : (s, a) \in \cS \times \cA \longmapsto
Q_{h}^{\bpi,\bcR_t}(s,a) - V_{h}^{\bpi,\bcR_t}(s)\,.
\end{equation}
We only keep in the notation $V_h$, $Q_h$, and $A_h$ the parameters $\bpi$ and $\bcR_t$ that vary, and omit the transition kernels $\bcT$.
We use the short-hand notation
\begin{equation}
\label{eq:def-A-bis}
A_h^{\bpi,\bcR_t}(s,\,\cdot\,) = \bigl( A_h^{\bpi,\bcR_t}(s,a) \bigr)_{a \in \cA}
\end{equation}
to denote the vector of advantages of a stationary policy $\bpi$ for a given episode~$t$ and a given stage~$h$.

\section{Methodology and core result: adversarial learning on advantage functions}
\label{sec:oracle}

\contrib{We recall how strategies designed to control the regret in the so-called adversarial setting,
i.e., satisfying guarantees as described in Definition~\ref{def:adv} below,
may be used to construct policies so as to control the regret
in terms of value functions. This observation was essentially already made
in the literature, at least for exponential weights; see, for instance, how \citet[Section~6]{Shani20}
handles the quantity called term~(ii) in their proof.}

Before formally stating our main result,
we briefly recall what the adversarial setting consists in, mostly to set our notation.
We assume that the reader is familiar with the fundamental concepts and results of
adversarial learning and refer to the monograph by \citet{CBL06} for a more detailed exposition.

\subsection{Reminder on adversarial learning}
\label{sec:reminder-advL}

We provide a description where $K \geq 2$ refers to the number of options that the learning
strategy has (the number of experts with the classic terminology of adversarial learning).
In Section~\ref{sec:advL-adv}, we will identify this set $[K]$ of options with the set of actions $\cA$.

At each round $t \geq 1$, based on the information collected during past rounds, a learning strategy picks
a convex combination $w_t = (w_{t,1},\ldots,w_{t,K}) \in \cP\bigl([K]\bigr)$ while an opponent
player simultaneously picks, possibly at random, a vector $g_t = (g_{t,1},\ldots,g_{t,K})$ of signed
rewards. Both $w_t$ and $g_t$ are revealed at the end of the round.
More formally, we mean that a learning strategy is
a sequence $\varphi = (\varphi_t)_{t \geq 1}$ of functions $\varphi_t : \R^{K(t-1)} \to \cP\bigl([K]\bigr)$
and that $w_t = \varphi_t\bigl( (g_\tau)_{\tau \leq t-1} \bigr)$ for $t \geq 1$.
This formula means in particular that the initial vector $w_1 = \varphi_1(\emptyset)$ is constant.

\begin{definition}[adversarial-learning regret bound]
\label{def:adv}
A sequential strategy controls the regret in the adversarial setting
with rewards bounded by $M > 0$
if there exists a sequence $(B_{T,K})_{T \geq 1}$ of positive numbers with $B_{T,K}/T \to 0$ and
such that, against all opponent players sequentially picking reward vectors in $[-M,M]^K$,
\[
\forall T \geq 1, \qquad
\max_{k \in [K]} \sum_{t=1}^T g_{t,k} - \sum_{t=1}^T \sum_{j \in [K]} w_{t,j}\,g_{t,j}
\leq 2M\,B_{T,K}\,.
\]
\end{definition}

The optimal orders of magnitude of $B_{T,K}$ are $\sqrt{T \ln K}$
(see \citealp{CBL06}).
In Definition~\ref{def:adv}, the strategy may know $M$ and rely on its value.
On the contrary, the number $T$ of rounds is unknown and in fact,
for the sake of exposition, Definition~\ref{def:adv} requires a control of the adversarial regret
for all $T \geq 1$, which imposes a mild restriction.

Two simple examples of strategies abiding by the constraints of
Definition~\ref{def:adv} are instances of the
potential-based strategies by \citet{CBL03}.
They are defined based on a sequence of non-decreasing functions $\Phi_t : \R \to [0,+\infty)$;
they resort to $w_{1,k} = 1/K$ and
\begin{equation}
\label{eq:def-pot-based}
\forall t \geq 2, \qquad
w_{t,k} = \frac{v_{t,k}}{\displaystyle{\sum_{j \in [K]} v_{t,j}}} \,, \qquad \mbox{where} \qquad
v_{t,k} = \Phi_t\!\left( \sum_{\tau=1}^{t-1} g_{\tau,k} - \sum_{\tau=1}^{t-1} \sum_{j \in [K]} w_{\tau,j} g_{\tau,j} \right).
\end{equation}

\begin{example}
\label{ex:pot}
\citet[Section~2]{CBL03} show that the strategy based on the constant
polynomial potentials $\Phi_t \equiv \Phi : x \mapsto \bigl( \max\{x,0\} \bigr)^{2 \ln K}$
provides the control $B_{T,K} = \sqrt{6 T \ln K}$ for
the regret in the adversarial setting.
\end{example}

\begin{example}
\label{ex:pot-exp-etat}
\citet{ACBG02} studied exponential potentials $\Phi_t(x) = \exp(\eta_t x)$
with time-varying learning rates $\eta_t = (1/M) \sqrt{(\ln K)/t}$.
This sequential strategy controls the regret with $B_{T,K} = \sqrt{T \ln K}$
in the adversarial setting.
\end{example}

A third example is of a different, not potential-based, nature.
\begin{example}
\label{ex:Zink}
The greedy projection algorithm of \citet{Zink03} relies
on a sequence $(\eta_t)_{t \geq 1}$ of positive step sizes
and sets $w_{t+1} = \proj(w_{t} + \eta_{t} \, g_{t})$
for $t \geq 1$, where $w_1 = (1/K,\ldots,1/K)$ and
where $\proj$ is the convex projection
onto $\cP\bigl([K]\bigr)$ in the Euclidean norm.
For the choices $\eta_t = (1/M) \sqrt{1/(2Kt)}$,
this strategy controls the regret in the adversarial setting
with $B_{T,K} = \sqrt{2KT}$.
\end{example}

Dozens of strategies satisfying the guarantees of Definition~\ref{def:adv}
exist.

\subsection{Policy optimization via adversarial learning on advantage functions}
\label{sec:advL-adv}

This section presents rather standard material and must be read accordingly.
Indeed, what follows is a reduction that was essentially known, though it has previously been applied only with exponential weights and on $Q$--values rather than advantage functions. The proof follows the one by \citet[Section~6]{Shani20}---see also \citet[proof of Theorem~16]{AKLM21}---, i.e.,
is based on the performance difference lemma.

We present the reduction in the ideal setting, where an oracle provides at the end of each episode $t$
the value functions of the policy $\bpi_t$ and of the reward function $\bcR_t$
selected by the learning system and the environment, respectively.
(The reward function $\bcR_t$ does not need to directly be revealed, though,
but only indirectly through the value functions.)
We consider this ideal setting throughout this article, except in
Section~\ref{sec:EstAdv}, where we explain how to leverage in practice the results
developed in the ideal setting.

\begin{oracle}
At the end of each episode $t \geq 1$, an oracle provides, for each $h \in [H]$,
the value functions
\[
Q_h^{\bpi_t,\bcR_t} : \cS \times \cA \to [0,H-h+1]\,, \quad
V_h^{\bpi_t,\bcR_t} : \cS \to [0,H-h+1]\,, \quad
A_h^{\bpi_t,\bcR_t} : \cS \times \cA \to \bigl[{-(H-h+1)},H-h+1\bigr]
\]
of the policy $\bpi_t$ and of the reward function $\bcR_t$
selected by the learning system and the environment, respectively.
\end{oracle}

For each stage $h \in [H]$,
we fix a sequential strategy $\varphi_h = (\varphi_{t,h})_{t \geq 1}$ in the adversarial setting,
relying on reward vectors bounded by $M_h = H-h+1$ and of dimension $K = A$, i.e., indexed by~$\cA$.
We run these strategies on the advantage functions, in a stage-by-stage and state-by-state manner,
as follows: for all $t \geq 1$,
\begin{equation}
\label{eq:ad-adv-def}
\forall h \in [H], \ \
\forall s \in \cS, \qquad
\pi_{t,h}(\,\cdot\, |s) = \varphi_{t,h} \Bigl( \bigl( A_h^{\bpi_\tau,\bcR_\tau}(s,\,\cdot\,) \bigr)_{\tau \leq t-1} \Bigr)\,,
\end{equation}
where we used the notation defined in~\eqref{eq:def-A-bis}. We refer to this strategy as
$(\varphi_h)_{h \in [H]}$--\texttt{Adv2}, for $(\varphi_h)_{h \in [H]}$--\underline{adv}ersarial learning on \underline{adv}antage functions.

It constitutes a ``theoretical'' strategy, as it relies on the oracle knowledge
of the advantage functions---an issue that we discuss and mitigate later in Section~\ref{sec:EstAdv}.
The strategy could be run instead on $Q$--values, see Remark~\ref{rk:A-Q-val} below.

\begin{theorem}
\label{th:main}
In the setting of Section~\ref{sec:setting-main} where rewards lie in $[0,1]$,
if, for all $h \in [H]$, the sequential strategies $\varphi_h$ control the regret
in the adversarial setting (Definition~\ref{def:adv}) by $B_{T,A}$ for $A$--dimensional
reward vectors bounded by $H-h+1$,
then the $(\varphi_h)_{h \in [H]}$--\texttt{\emph{Adv2}} strategy defined in~\eqref{eq:ad-adv-def}
controls the regret as:
\[
\forall T \geq 1, \qquad
\max_{\bpi} \sum_{t=1}^T \Bigl( V_1^{\bpi,\bcR_t}(s_1) - V_1^{\bpi_t,\bcR_t}(s_1) \Bigr)
\leq H(H+1)\,B_{T,A}\,.
\]
\end{theorem}

As indicated above,
following \citet[Section~6]{Shani20}, the (short) proof of Theorem~\ref{th:main} relies on the
so-called performance difference lemma, which we recall next.
For the sake of completeness, references for this lemma and a proof thereof are provided
in Appendix~\ref{sec:PDL}.

\begin{restatable}[Performance difference lemma]{lemma}{lmPDL}
\label{lm:PDL}
Let $\mu_{h'}^{s_1,\bpi,\bcT}$ be the distribution of the state $s_{h'}$
of the $h'$--th stage, starting from the state $s_1$ in the first stage,
following the stationary policy $\bpi$ and the transition kernels $\bcT$.
In a MDP with transition kernels $\bcT$,
for all pairs $\bpi,\bpi'$ of stationary policies, for all reward functions $\bcR$,
and for all stages $h \in [H]$,
\[
\sum_{s \in \cS} \mu_{h}^{s_1,\bpi,\bcT}(s) \Bigl( V_h^{\bpi,\bcR}(s) - V_h^{\bpi',\bcR}(s) \Bigr)
= \sum_{h'=h}^H \sum_{s \in \cS} \mu_{h'}^{s_1,\bpi,\bcT}(s) \sum_{a \in \cA} \pi_{h'}(a|s) \, A^{\bpi',\bcR}_{h'}(s,a)\,.
\]
In particular, for $h=1$,
\[
V_1^{\bpi,\bcR}(s_1) - V_1^{\bpi',\bcR}(s_1)
= \sum_{h'=1}^H \sum_{s \in \cS} \mu_{h'}^{s_1,\bpi,\bcT}(s) \sum_{a \in \cA} \pi_{h'}(a|s) \, A^{\bpi',\bcR}_{h'}(s,a)\,.
\]
\end{restatable}

\begin{proof}[Proof of Theorem~\ref{th:main}]
We fix a stationary policy $\bpi$ throughout the proof and control the regret with respect to this~$\bpi$.

The first part consists of applying the adversarial-learning regret upper bound for each $h \in [H]$.
As the rewards take values in~$[0,1]$, we have that $\bigl| A_h^{\bpi_\tau,\bcR_\tau}(s,a) \bigr| \leq H-h+1$
for all $\tau,s,a$.
By the definition of advantage functions (for the equality to~$0$) and by
Definition~\ref{def:adv} and the design of the $(\varphi_h)_{h \in [H]}$--\texttt{Adv2} strategy (for the upper bound),
we have, for all $s \in \cS$,
\begin{equation}
\label{eq:csq-agreg-A}
\max_{a \in \cA} \sum_{t=1}^T A^{\bpi_t,\bcR_t}_{h}(s,a)
- \sum_{t=1}^T \overbrace{\sum_{a \in \cA} \pi_{t,h}(a|s) \, A^{\bpi_t,\bcR_t}_{h}(s,a)}^{=\,0}
\leq 2(H-h+1)\,B_{T,A}\,.
\end{equation}
The second part consists of applying
the performance difference lemma, i.e., Lemma~\ref{lm:PDL} above with $h=1$, which guarantees that
\[
V_1^{\bpi,\bcR_t}(s_1) - V_1^{\bpi_t,\bcR_t}(s_1)
= \sum_{h=1}^H \sum_{s \in \cS} \mu_h^{s_1,\bpi,\bcT}(s) \sum_{a \in \cA} \pi_h(a|s) \, A^{\bpi_t,\bcR_t}_{h}(s,a)\,.
\]
Summing this equality over $t$ and rearranging, we get
\begin{align}
\nonumber
\sum_{t=1}^T \Bigl( V_1^{\bpi,\bcR_t}(s_1) - V_1^{\bpi_t,\bcR_t}(s_1) \Bigr)
& = \sum_{h=1}^H \sum_{s \in \cS} \mu_h^{s_1,\bpi,\bcT}(s) \sum_{a \in \cA} \pi_h(a|s) \sum_{t=1}^T A^{\bpi_t,\bcR_t}_{h}(s,a) \\
\label{eq:PDL-main}
& \leq \sum_{h=1}^H \sum_{s \in \cS} \mu_h^{s_1,\bpi,\bcT}(s) \,\, \underbrace{\max_{a \in \cA} \sum_{t=1}^T A^{\bpi_t,\bcR_t}_{h}(s,a)}_{\leq
2 (H-h+1)\,B_{T,A}} \ \leq \underbrace{2 \sum_{h=1}^H (H-h+1)}_{= H(H+1)} \, B_{T,A} \,,
\end{align}
where we substituted~\eqref{eq:csq-agreg-A}. Here, we crucially used that the weights
$\mu_h^{s_1,\bpi,\bcT}(s)$ are independent of $t$ as they only depend on the fixed benchmark policy~$\bpi$,
on the common transition kernels $\bcT$, and on the initial state~$s_1$ (identical for all $t$).
\end{proof}

\subsection{Comments}
\label{sec:comments-Q}

In this section, we comment and discuss the \texttt{Adv2} strategy~\eqref{eq:ad-adv-def}
and its bound.

We first note that the regret bound of Theorem~\ref{th:main} is independent of the size $S$ of the state space;
it only depends on the size $A$ of the action space, on the number $T$ of episodes, and on the length $H$ of the episodes.
Given that adversarial-learning strategies have a per-round computational complexity typically proportional
to~$K$ (with the notation of Section~\ref{sec:reminder-advL}),
the per-round computational complexity of the \texttt{Adv2} strategies~\eqref{eq:ad-adv-def}
are typically proportional to $SAH$ as far as the weight updates are concerned.
The main computational issue lies in computing (or estimating, see Section~\ref{sec:EstAdv})
the advantage functions $A_h^{\bpi_\tau,\bcR_\tau}$.

Second, for potential-based strategies~\eqref{eq:def-pot-based},
we note that the original definition~\eqref{eq:ad-adv-def} of \texttt{Adv2}
and the alternative definition based on $Q$--values,
\begin{equation}
\label{eq:ad-adv-def-Q}
\pi_{t,h}(\,\cdot\, |s) = \varphi_{t,h} \Bigl( \bigl( Q_h^{\bpi_\tau,\bcR_\tau}(s,\,\cdot\,) \bigr)_{\tau \leq t-1} \Bigr)\,,
\end{equation}
lead to the exact same strategies.
This may be shown by induction, based on the fact that for all $h \in [H]$ and $(s,a) \in \cS \times \cA$,
as in~\eqref{eq:csq-agreg-A} for the first equality and due to the definitions of value
functions for the second equality,
\begin{align*}
& \sum_{\tau = 1}^{t-1} A_h^{\bpi_\tau,\bcR_\tau}(s,a) -
\sum_{\tau = 1}^{t-1} \overbrace{\sum_{a \in \cA} \pi_{\tau,h}(a|s) \, A^{\bpi_\tau,\bcR_\tau}_{h}(s,a)}^{= 0}
= \sum_{\tau = 1}^{t-1} A_h^{\bpi_\tau,\bcR_\tau}(s,a) \\
\mbox{and} \qquad &
\sum_{\tau = 1}^{t-1} Q_h^{\bpi_\tau,\bcR_\tau}(s,a) -
\sum_{\tau = 1}^{t-1} \underbrace{\sum_{a \in \cA} \pi_{\tau,h}(a|s) \, Q^{\bpi_\tau,\bcR_\tau}_{h}(s,a)}_{=
V_h^{\bpi_\tau,\bcR_\tau}(s)}
= \sum_{\tau = 1}^{t-1} A_h^{\bpi_\tau,\bcR_\tau}(s,a)\,.
\end{align*}

For general adversarial-learning strategies, the induced strategies~\eqref{eq:ad-adv-def} and~\eqref{eq:ad-adv-def-Q} may differ,
though they achieve the same regret guarantees, as detailed by the following remark.

\begin{remark}
\label{rk:A-Q-val}
An inspection of the proof of Theorem~\ref{th:main} shows that it would also work for the strategies
of the form~\eqref{eq:ad-adv-def-Q}.
Indeed, the inequality~\eqref{eq:csq-agreg-A} therein would be replaced equivalently by
\[
2(H-h+1)\,B_{T,A} \geq
\max_{a \in \cA} \sum_{t=1}^T Q^{\bpi_t,\bcR_t}_{h}(s,a)
- \sum_{t=1}^T \overbrace{\sum_{a \in \cA} \pi_{t,h}(a|s) \, Q^{\bpi_t,\bcR_t}_{h}(s,a)}^{= V^{\bpi_t,\bcR_t}_{h}(s)}
= \max_{a \in \cA} \sum_{t=1}^T A^{\bpi_t,\bcR_t}_{h}(s,a)\,,
\]
while the rest of the proof would be unaffected. However, using the advantage functions
is preferred in practice, as it provides a greater numerical stability, as well as
a possibly lower variance when the value function are estimated (see Section~\ref{sec:EstAdv}).
\end{remark}

\section{Extension 1: \\ ~~~~\,Convergence of the last iterate for some adversarial learning strategies}
\label{sec:monot}

\contrib{We generalize an argument of \citet[Section~5.3]{AKLM21},
which was provided for exponential weights only (in the discounted setting): the
aim is to control the convergence of the last iterate, i.e., to upper bound
$\smash{\displaystyle{\max_{\bpi} V_1^{\bpi,\bcR}(s_1) - V_1^{\bpi_T,\bcR}(s_1)}}$,
when (mean) rewards functions are constant over time. \smallskip \newline
To do so, we introduce
a concept of independent interest: monotonicity of weights for adversarial-learning
strategies.}

More precisely, for adversarial-learning strategies $\varphi$ satisfying this property of
monotonicity of weights,
and in case reward functions do not vary over time
(or even just mean reward functions do not vary over time, see Remark~\ref{rk:mrf})
the result of Theorem~\ref{th:main} may be strengthened into a convergence result
of the last iterate, at a rate faster by a factor of $1/T$ compared to
the convergence of the cumulative regret~\eqref{eq:def-regret-V}.

\begin{definition}[monotonicity of weights]
\label{def:monot}
A sequential strategy $\varphi = (\varphi_t)_{t \geq 1}$ in the adversarial setting satisfies monotonicity of weights
if against all opponent players sequentially picking $K$--dimensional reward vectors $g_\tau = (g_{\tau,k})_{k \in [K]}$,
the convex weights output by $\varphi$ are such that
\[
\forall t \geq 1, \qquad
\sum_{k \in [K]} w_{t+1,k} \left( g_{t,k} - \sum_{j \in [K]} w_{t,j} g_{t,j} \right) \geq 0\,,
\]
where we recall the notation
$(w_{t,k})_{k \in [K]} = \varphi_t(g_1,\ldots,g_{t-1})$ and $(w_{t+1,k})_{k \in [K]} = \varphi_t(g_1,\ldots,g_{t-1},g_t)$.
\end{definition}

The proof of Lemma~\ref{lm:monot} below explains why the property of
Definition~\ref{def:monot} is termed monotonicity of weights, and why it is a natural property
of an adversarial learning strategy:
indeed, the property
is satisfied as soon as weights for components $k$
associated with a good (respectively, bad) reward $g_{t,k}$ in the previous round increase (respectively,
decrease), where good or bad is determined by the sign of what is called the instantaneous regret
with respect to component $k$ in round $t$:
\[
g_{t,k} - \sum_{j \in [K]} w_{t,j} g_{t,j}\,.
\]

\begin{restatable}{lemma}{lmMonot}
\label{lm:monot}
The potential-based strategies~\eqref{eq:def-pot-based} of~\citet{CBL03} with constant, non-decreasing potential
functions $\Phi_t \equiv \Phi$ (like in Example~\ref{ex:pot})
and the greedy projection algorithm (Example~\ref{ex:Zink}) of~\citet{Zink03}
satisfy monotonicity of weights.
\end{restatable}

\begin{proof}
We start with the potential-based strategies~\eqref{eq:def-pot-based}, in case of a constant,
non-decreasing potential function $\Phi_t \equiv \Phi$, and use the notation
defined therein. For each $t \geq 1$, since $\Phi$ is non-decreasing, we have,
for all $k \in [K]$,
\[
v_{t+1,k} \geq v_{t,k} \quad \Longleftrightarrow \quad g_{t,k} - \sum_{j \in [K]} w_{t,j} g_{t,j} \geq 0\,,
\qquad \mbox{thus} \qquad
(v_{t+1,k} - v_{t,k}) \left( g_{t,k} - \sum_{j \in [K]} w_{t,j} g_{t,j} \right) \geq 0
\]
in all cases. Therefore,
\[
\sum_{k \in [K]} v_{t+1,k} \left( g_{t,k} - \sum_{j \in [K]} w_{t,j} g_{t,j} \right)
\geq \sum_{k \in [K]} v_{t,k} \left( g_{t,k} - \sum_{j \in [K]} w_{t,j} g_{t,j} \right) = 0\,,
\]
where the equality to~$0$ and the final result of Definition~\ref{def:monot}
are obtained, respectively, by normalizing the $v_{t+1,k}$ and $v_{t,k}$
into $w_{t+1,k}$ and $w_{t,k}$.

For the greedy projection algorithm (Example~\ref{ex:Zink}) of~\citet{Zink03},
we note that by a property of Euclidean projections onto a convex set
(here, $w_{t+1}$ is the projection of $w_t + \eta_t \, g_t$ onto
the simplex, and $w_t$ also belongs to the simplex),
the following Euclidean inner product is non-positive:
\[
0 \geq \bigl\langle w_t - w_{t+1}, \,\, (w_t + \eta_t \, g_t) - w_{t+1} \bigr\rangle
= \Arrowvert w_t - w_{t+1} \Arrowvert^2 + \eta_t \langle w_t - w_{t+1}, \, g_t \rangle\,,
\]
so that $\langle w_{t+1}-w_t, \, g_t \rangle \geq 0$,
which is exactly monotonicity of weights.
\end{proof}

We are now ready to state our result of convergence
of the last iterate, which generalizes an argument of \citet[Section~5.3]{AKLM21}.

\begin{theorem}
\label{th:main-bis}
Assume reward functions do not vary over time and are all equal to some~$\bcR$.
If, for all $h \in [H]$, the sequential strategies $\varphi_h$ satisfy monotonicity of weights (Definition~\ref{def:monot})
and control the regret in the adversarial setting (Definition~\ref{def:adv}) by $B_{T,A}$ for $A$--dimensional
reward vectors bounded by $H-h+1$,
then the last iterate of the $(\varphi_h)_{h \in [H]}$--\texttt{\emph{Adv2}} strategy defined in~\eqref{eq:ad-adv-def}
satisfies
\[
\forall T \geq 1, \qquad
\max_{\bpi} V_1^{\bpi,\bcR}(s_1) - V_1^{\bpi_T,\bcR}(s_1)
\leq \frac{H(H+1)\,B_{T,A}}{T}\,.
\]
\end{theorem}

The bound by~\citet[Section~5.3]{AKLM21}, where the exponential weights with a constant learning rate
are considered, corresponds to this theorem
but is stated separately in Corollary~\ref{cor:NPG},
for reasons that will be made clear in Section~\ref{sec:NPG}.
As the proof of Theorem~\ref{th:main-bis} is concise, we provide it in the main body
of this article.

\begin{proof}
Given the definition~\eqref{eq:ad-adv-def}, the monotonicity of weights (Definition~\ref{def:monot}),
and the definition of advantage functions,
we have that, for all $t \geq 1$, for all $h \in [H]$, and $s \in \cS$,
\[
\sum_{a \in \cA} \pi_{t+1,h}(a|s) \, A^{\bpi_t,\bcR}_{h}(s,a) \geq
\sum_{a \in \cA} \pi_{t,h}(a|s) \, A^{\bpi_t,\bcR}_{h}(s,a) = 0 \,.
\]
Therefore, the performance difference lemma, i.e., Lemma~\ref{lm:PDL} above with $h=1$, shows that
\[
V_1^{\bpi_{t+1},\bcR}(s_1) - V_1^{\bpi_t,\bcR}(s_1)
= \sum_{h=1}^H \sum_{s \in \cS} \mu_h^{s_1,\bpi_{t+1},\bcT}(s)
\underbrace{\sum_{a \in \cA} \pi_{t+1,h}(a|s) \, A^{\bpi_t,\bcR}_{h}(s,a)}_{\geq \, 0} \geq 0
\,.
\]
(This is the part of the proof where we crucially use that reward functions do not vary over time.)
Thus,
\[
\max_{\bpi} V_1^{\bpi,\bcR}(s_1) - V_1^{\bpi_T,\bcR}(s_1)
\leq \max_{\bpi} V_1^{\bpi,\bcR}(s_1) - \frac{1}{T} \sum_{t=1}^T V_1^{\bpi_t,\bcR}(s_1)
\leq \frac{H(H+1)\,B_{T,A}}{T}\,,
\]
where we applied Theorem~\ref{th:main} for the final bound.
\end{proof}

\begin{remark}
\label{rk:mrf}
An inspection of the proof above shows that what actually matters is
only that \emph{mean} reward functions $\br_t = (r_{t,h})_{h \in [H]}$ be constant over time.
Indeed, the value and advantage functions only depend on the $\bcR_t$ through the $\br_t$;
this fact is also illustrated in the proof of the performance difference lemma
which only requires identical mean reward functions, not the identity of reward functions.
\end{remark}

\section{Extension 2: Stronger forms of regret}
\label{sec:stronger-regret}

\contrib{We push the logic of the reduction of the control of MDPs to adversarial learning,
and leverage stronger forms of regret in adversarial learning. This section
thus presents new regret criteria for learning MDPs.}

Definition~\ref{def:adv} considers the simplest definition of adversarial regret.
However, several stronger notions of regrets were proposed in the literature.
The proof of Theorem~\ref{th:main} shows that the vanilla notion of adversarial regret
of Definition~\ref{def:adv}
may be transferred into the vanilla regret~\eqref{eq:def-regret-V} in terms of value functions.
Actually, this proof may be mimicked to transfer stronger notions of adversarial regret.
We illustrate this possibility with two notions of adversarial regrets that replace the comparison to a
single global policy with local comparisons (strongly adaptive regret)
or by global comparisons to sequences of policies (tracking regret).

\subsection{Strongly adaptive regret and tracking regret in adversarial learning}

We use again the notation for adversarial learning introduced at the beginning of Section~\ref{sec:direct}.
The first extended notion of regret, called strongly adaptive regret, measures performance
simultaneously over each given sub-interval of time with respect to the best component over that sub-interval.
It was introduced by~\citet{DGSSS15}, based on the concept of adaptive regret from~\citet{HS09},
itself based on the work by~\citet{LS94}.

\begin{definition}[strongly adaptive regret in adversarial learning]
\label{def:SAR}
A sequential strategy controls the strongly adaptive regret in the adversarial setting
with rewards bounded by $M > 0$
if there exist positive numbers $B_{T,K,\tau}$, where $T \geq 1$ and $\tau \in [T]$,
such that, against all opponent players sequentially picking reward vectors in $[-M,M]^K$,
\[
\forall T \geq 1, \quad \forall \tau \in [T], \qquad
\max_{t_0 \in [T-\tau+1]} \left\{
\max_{k \in [K]} \sum_{t=t_0}^{t_0+\tau-1} g_{t,k} - \sum_{t=t_0}^{t_0+\tau-1} \sum_{j \in [K]} w_{t,j}\,g_{t,j} \right\}
\leq 2M\,B_{T,K,\tau}\,,
\]
and \quad $\displaystyle{\sup_{\tau \in [T]} \frac{B_{T,K,\tau}}{T}} \to 0$ \quad as $T \to \infty$.
\end{definition}

It follows from \citet[Theorem~1]{DGSSS15} that the strongly adaptive regret can be controlled with bounds
$B_{T,K,\tau}$ of order $\sqrt{\tau}$ up to logarithmic factors.

A closely related notion is the tracking regret,
introduced by~\citet{HW98} (see also \citealp[Chapter~5.2]{CBL06}),
where the comparison is taken over all time steps but against sequences
$k_{1:T} = (k_1,\,k_2,\,\ldots,\,k_T)$ with values in~$[K]$,
and containing at most $C$ shifts (i.e., $C$ time steps such that $k_t \ne k_{t-1}$).
The tracking regret involves
\[
\sum_{t=1}^T g_{t,k_t} - \sum_{t=1}^T \sum_{j \in [K]} w_{t,j}\,g_{t,j}\,.
\]
There are strong links between strongly adaptive and tracking regret, see~\citet{AKCV16}.
In particular, we explain, in the context of regret with value functions,
how strongly adaptive regret with $B_{T,K,\tau}$ of order $\sqrt{\tau}$ up to logarithmic factors
entails tracking regret of order $\sqrt{CT}$; see Corollary~\ref{cor:stronger}.

\subsection{Transfer of strongly adaptive regret bounds}

Based on Definition~\ref{def:SAR},
we obtain the following regret bound in terms of value functions
and policies.

\begin{theorem}
\label{th:stronger}
In the setting of Section~\ref{sec:setting-main} where rewards lie in $[0,1]$,
if, for all $h \in [H]$, the sequential strategies $\varphi_h$ control the strongly adaptive regret
in the adversarial setting (Definition~\ref{def:SAR}) by $B_{T,A,\tau}$ for $A$--dimensional
reward vectors bounded by $H-h+1$,
then the $(\varphi_h)_{h \in [H]}$--\texttt{\emph{Adv2}} strategy defined in~\eqref{eq:ad-adv-def}
ensures that
\[
\forall T \geq 1, \quad \forall \tau \in [T], \qquad
\max_{t_0 \in [T-\tau+1]} \left\{
\max_{\bpi} \sum_{t=t_0}^{t_0+\tau-1} \Bigl( V_1^{\bpi,\bcR_t}(s_1) - V_1^{\bpi_t,\bcR_t}(s_1) \Bigr) \right\}
\leq H(H+1)\,B_{T,A,\tau}\,.
\]
\end{theorem}

The proof of Theorem~\ref{th:stronger} is obtained by a direct adaptation of the proof of Theorem~\ref{th:main},
which basically consists of considering sums
over sub-intervals only instead of sums over all time periods. Again, since
the proof is concise, we provide it here.

\begin{proof}[Proof of Theorem~\ref{th:stronger}]
We fix a stationary policy $\bpi$ throughout the proof and control some adaptive regret with respect to this~$\bpi$.
By the design~\eqref{eq:ad-adv-def} of the \texttt{Adv2} strategy,
which operates stage by stage and state by state, we have that for all $h \in [H]$ and $s \in \cS$,
the following holds, by Definition~\ref{def:SAR}:
for all $T \geq 1$ and $\tau \in [T]$,
\[
\max_{t_0 \in [T-\tau+1]} \left\{
\max_{a \in \cA} \sum_{t=t_0}^{t_0+\tau-1} A^{\bpi_t,\bcR_t}_{h}(s,a)
- \sum_{t=t_0}^{t_0+\tau-1} \overbrace{\sum_{a \in \cA} \pi_{t,h}(a|s) \, A^{\bpi_t,\bcR_t}_{h}(s,a)}^{=\,0}
\right\} \leq 2(H-h+1)\,B_{T,A,\tau}\,.
\]
The same application of the performance difference lemma as in the proof of Theorem~\ref{th:main}
entails that for all $T \geq 1$, $\tau \in [T]$, and $t_0 \in [T-\tau+1]$,
\begin{align*}
\sum_{t=t_0}^{t_0+\tau-1} \Bigl( V_1^{\bpi,\bcR_t}(s_1) - V_1^{\bpi_t,\bcR_t}(s_1) \Bigr)
& = \sum_{h=1}^H \sum_{s \in \cS} \mu_h^{s_1,\bpi,\bcT}(s) \sum_{a \in \cA} \pi_h(a|s) \sum_{t=t_0}^{t_0+\tau-1} A^{\bpi_t,\bcR_t}_{h}(s,a) \\
& \leq \sum_{h=1}^H \sum_{s \in \cS} \mu_h^{s_1,\bpi,\bcT}(s) \,\, \underbrace{\max_{a \in \cA} \sum_{t=t_0}^{t_0+\tau-1} A^{\bpi_t,\bcR_t}_{h}(s,a)}_{\leq
2 (H-h+1)\,B_{T,A,\tau}} \ \leq H(H+1) \, B_{T,A,\tau} \,.
\end{align*}
Here again, we crucially used that the weights $\mu_h^{s_1,\bpi,\bcT}(s)$ are independent of $t$.
The claimed bound follows by taking the maximum over $\bpi$ and over $t_0 \in [T-\tau+1]$.
\end{proof}

\subsection{Tracking regret bounds}

We detail a consequence of the bound of Theorem~\ref{th:stronger} in terms of tracking regret.

We now consider sequences $\bpi^{(1:T)} = \bigl( \bpi^{(1)}, \, \bpi^{(2)}, \, \ldots, \, \bpi^{(T)} \bigr)$
of stationary policies as comparison points, instead of a single stationary policy.
We define the number of shifts $\smash{c\bigl( \bpi^{(1:T)} \bigr)}$ of such a sequence
as follows: the smallest integer $c'$ such that
there exist $c'-1$ integers $\tau_2,\ldots,\tau_{c'}$ with values in $[T]$ such that,
denoting $\tau_1 = 1$ and $\tau_{c'+1} = T+1$,
\[
\forall i \in \{2,\ldots,c'+1\}, \qquad \forall t \in \{ \tau_{i-1}, \ldots, \tau_i - 1 \}, \qquad
\bpi^{(t)} = \bpi^{(\tau_{i-1})}\,.
\]
The tracking regret against sequences $\bpi^{(1:T)}$ of stationary policies with at most $C$ shifts
is defined as
\[
\underset{\substack{\bpi^{(1:T)} \ \mbox{\tiny \rm such that} \\ c(\bpi^{(1:T)}) \leq C}}{\max} \,\,
\sum_{t=1}^T V_1^{\bpi^{(t)},\bcR_t}(s_1) - \sum_{t=1}^T V_1^{\bpi_t,\bcR_t}(s_1)\,.
\]
We fix $1 \leq C \leq T$ and a sequence $\bpi^{(1:T)}$ of stationary policies, with at most $C$ shifts,
occurring at episodes $1 = \tau_1 \leq \tau_2 \leq \ldots \leq \tau_C$.
(The inequalities are strict if there are exactly $C$ shifts.)
We introduce $\tau_{C+1} = T+1$ and partition time into the $C$ intervals
$[\tau_i, \, \tau_{i+1}-1]$, for $i \in [C]$.
The values successively
taken by the sequence $\bpi^{(1:T)}$ consist of the $\bpi^{(\tau_i)}$, where $i \in [C]$.
By applying the bound of Theorem~\ref{th:stronger} on each of the $C$ intervals
$[\tau_i, \, \tau_{i+1}-1]$, we obtain the following corollary.

\begin{corollary}
\label{cor:stronger}
Under the assumptions of Theorem~\ref{th:stronger},
the $(\varphi_h)_{h \in [H]}$--\texttt{\emph{Adv2}} strategy defined in~\eqref{eq:ad-adv-def}
also ensures that \qquad $\forall T \geq 1, \quad \forall C \in [T],$
\[
\underset{\substack{\bpi^{(1:T)} \ \mbox{\tiny \rm such that} \\ c(\bpi^{(1:T)}) \leq C}}{\max} \,\,
\sum_{t=1}^T V_1^{\bpi^{(t)},\bcR_t}(s_1) - \sum_{t=1}^T V_1^{\bpi_t,\bcR_t}(s_1)
\leq H(H+1) \, \max_{\substack{1 = \tau_1 \leq \tau_2 \leq \ldots \\ \ \ \leq \tau_C \leq \tau_{C+1} = T+1}} \sum_{i=1}^C B_{T,A,\tau_{i+1}-\tau_i}\,.
\]
\end{corollary}

In particular, if $B_{T,A,\tau} \leq \ell(T,K) \, \sqrt{\tau}$, where $\ell(T,K)$ is logarithmic in $T$ and $K$, which is
a standard bound, then
by Jensen's inequality for $\sqrt{\,\cdot\,}$,
\[
\max_{\substack{1 = \tau_1 \leq \tau_2 \leq \ldots \\ \ \ \leq \tau_C \leq \tau_{C+1} = T+1}} \sum_{i=1}^C B_{T,A,\tau_{i+1}-\tau_i}
\leq \ell(T,K) \, \max_{\substack{1 = \tau_1 \leq \tau_2 \leq \ldots \\ \ \ \leq \tau_C \leq \tau_{C+1} = T+1}} \underbrace{\sum_{i=1}^C \sqrt{\tau_{i+1}-\tau_i}}_{\leq \sqrt{C (\tau_{C+1} - \tau_1)} = \sqrt{CT}}
\leq \ell(T,K) \, \sqrt{C T}\,.
\]

\section{The special case of exponential weights: improved regret bounds}
\label{sec:NPG}

\contrib{The literature (see Section~\ref{sec:lit-review}) essentially focuses on the adversarial learning strategy
given by exponential weights with a constant learning rate~$\eta$. It turns out that this strategy
does not satisfy the requirement of Definition~\ref{def:adv} because of a tuning issue: the adversarial regret
bound is of the form $\ln N/\eta + \eta M T/2$ (see, e.g., \citealp[Theorem~2.2]{CBL06}) and cannot be simultaneously optimized for all
values of $T$. The literature typically assumes that $T$ is known and obtains a $\sqrt{T}$
regret bound for MDPs by taking $\eta$ of order $1/\sqrt{T}$; see, for instance,
among many others, \citet{Cai20} and \citet{Shani20}. A notable exception,
in the discounted setting and for a constant reward function, can be extracted from the proof of
\citet[Section~5.3]{AKLM21}---they handle convergence of the last iterate but their proof technique
also applies to cumulative regret. We extend their result to the episodic setting
and show that it is not essential that the reward functions be constant over time:
we provide an upper bound in terms of the numbers of shifts in the sequence of reward functions.}

We study in this section the strategy~\eqref{eq:ad-adv-def} of Section~\ref{sec:advL-adv}
where the adversarial learning strategies are given by the
strategy~\eqref{eq:def-pot-based} based on a constant exponential potential $\Phi_t \equiv \Phi : x \mapsto \exp( \eta x)$.
This strategy takes the following simple form: for all $t \geq 1$,
\begin{equation}
\label{eq:ad-adv-def-EXP}
\forall h \in [H], \ \
\forall s \in \cS, \ \
\forall a \in \cA, \qquad
\pi_{t,h}(a|s) = \frac{\displaystyle{\exp \!\left( \eta \sum_{\tau=1}^{t-1} A_h^{\bpi_\tau,\bcR_\tau}(s,a) \right)}}{
\displaystyle{\sum_{a' \in \cA} \exp \!\left( \eta \sum_{\tau=1}^{t-1} A_h^{\bpi_\tau,\bcR_\tau}(s,a') \right)}}\,,
\end{equation}
with the understanding that a sum over no term is null, i.e., $\pi_{1,h}(a|s) = 1/A$.

\citet[Section~5.3]{AKLM21} showed that the strategy above corresponds to the
natural policy gradient [NPG] strategy based on a softmax parametrization.
They proposed a direct analysis (in the discounted setting) with reward functions constant over time.
We adapt and extend this analysis to (obliviously) adversarial sequences of reward functions.
We also claim a more transparent proof scheme, consisting
of a suitable adversarial bound (finer than the uniform bounds considered in Definition~\ref{def:adv},
which in this case would be linear in $T$, as recalled in the introduction of this section)
applied to policy learning along the lines of the proof of Theorem~\ref{th:main}.

Our result is stated in terms of the number $R$ of regimes shifts in the sequence $\bcR_1,\ldots,\bcR_T$
of payoff functions. More formally, $R$ is the smallest integer such that
there exist $R-1$ integers $\tau_2,\ldots,\tau_{R}$ with values in $[T]$ such that,
denoting $\tau_1 = 1$ and $\tau_{R+1} = T+1$,
\begin{equation}
\label{eq:def-R}
\forall k \in \{2,\ldots,R+1\}, \qquad \forall t \in \{ \tau_{k-1}, \ldots, \tau_k - 1 \}, \qquad
\bcR_t = \bcR_{\tau_{k-1}}\,.
\end{equation}
(The case $R=1$ corresponds to a single regime, i.e., the reward functions $\bcR_t$
are independent of time.)

The proof of Theorem~\ref{th:NPG} below may be found in Appendix~\ref{app:NPG}.
It is more complex than the proof by \citet[Section~5.3]{AKLM21},
which could use a simple argument specific to the discounted setting, with discount factor~$\gamma$:
that distributions over states induced by a starting state $s_0$, a policy, and a transition function,
put a probability mass at least $1-\gamma$ on $s_0$, no matter the policy and the transition function.
See Remark~\ref{rk:proof-diff} for more details.

\begin{restatable}{theorem}{thNPG}
\label{th:NPG}
In the setting of Section~\ref{sec:setting-main} where rewards lie in $[0,1]$,
the policy learning strategy~\eqref{eq:ad-adv-def-EXP}
controls the regret as
\[
\max_{\bpi} \sum_{t=1}^T \Bigl( V_1^{\bpi,\bcR_t}(s_1) - V_1^{\bpi_t,\bcR_t}(s_1) \Bigr)
\leq \frac{H \ln A}{\eta} + R \, \frac{H (H+1)}{2}\,,
\]
where $R$ is the number of regime shifts in the sequence $\bcR_1,\ldots,\bcR_T$
of payoff functions.
\end{restatable}

The bound of Theorem~\ref{th:NPG} has a smaller order of magnitude
than the one of Theorem~\ref{th:main}, which is typically of order $\sqrt{T}$,
as soon as the number of regime shifts satisfies $R \ll \sqrt{T}$.
(In general, up to $T-1$ regime shifts may occur.)
In particular, the regret upper bound of Theorem~\ref{th:NPG} is smaller than a constant when
the reward functions do not vary over time.
Of course, as already mentioned at the beginning of Section~\ref{sec:advL-adv},
this observation is somewhat secondary in the absence of an oracle for value functions,
when value functions have to be estimated and when these estimation errors
are the main contributors to the regret bounds; see Section~\ref{sec:EstAdv}.

By Lemma~\ref{lm:monot} and (the proof of) Theorem~\ref{th:main-bis},
we have the following corollary to Theorem~\ref{th:NPG},
in case of a constant sequence of payoff functions.
It corresponds to the bound of \citet[Section~5.3]{AKLM21}
with $H$ playing the role of $1/(1-\gamma)$ therein.

\begin{corollary}
\label{cor:NPG}
In the setting of Section~\ref{sec:setting-main} where rewards lie in $[0,1]$,
if the reward functions do not vary over time and are all equal to some $\bcR$,
then the last iterate of the policy learning strategy~\eqref{eq:ad-adv-def-EXP}
satisfies
\[
\max_{\bpi} V_1^{\bpi,\bcR}(s_1) - V_1^{\bpi_T,\bcR}(s_1)
\leq \frac{H \ln A}{\eta T} + \frac{H (H+1)}{2T}\,.
\]
\end{corollary}

As in \citet[Section~5.3]{AKLM21}, the bounds obtained in
Theorem~\ref{th:NPG} and Corollary~\ref{cor:NPG} suggest choosing $\eta$
as large as possible. While this is the choice recommended by theory, practical performance
may be affected: in Section~\ref{sec:EstAdv}, we recommend to conduct empirical evaluations
to investigate this issue.

\section{Extension 3: \\ ~~~~\,Aggregation (orchestration) of expert policies}
\label{sec:aggreg}

\contrib{Adversarial learning is sometimes called prediction with experts (see \citealp{CBL06}).
We further pursue the idea of the reduction of the control of MDPs to adversarial learning
and now rather aggregate expert policies.
The aim is to mimic the performance of the overall best
convex combination of expert policies (which is, in particular,
better than the performance of the best policy taken in isolation),
which corresponds to an aggregation (or orchestration) of expert policies.
This setting was also termed
learning from multiple oracles (which may be understood as a specific paradigm in the vast imitation-learning literature) by
\citet{CKA20} and \citet{Liu23}. We obtain stronger forms of performance guarantees
than in the latter references, see Remark~\ref{rk:imitguar}.
We do so via some reduction to the standard tabular case
for a lifted MDP.}

We return to the considerations of Section~\ref{sec:direct} and
consider a finite number $K$ of stationary policies.
We denote by $\Pi = \{ \bpi_1,\ldots,\bpi_K \}$ the set of these policies and refer to
them as expert policies.
Furthermore, for a given stage~$h \in [H]$, we denote by $\Pi_h = \{ \pi_{1,h},\,\ldots,\,\pi_{K,h} \}$
the set of the corresponding policies.

We combine expert policies over time through
state-stage-dependent weights $\bp_t = (p_{t,h})_{h \in [H]} \in \cP\bigl([K]\bigr)^{[H] \times \cS}$,
where $p_{t,h}(\,\cdot\,|s) \in \cP\bigl([K]\bigr)$
may be interpreted either as a probability distribution over the policies in $\Pi_{h}$
or as providing convex weights for the aggregation of the policies in $\Pi_{h}$.
More precisely, for each episode $t \geq 1$, we denote by $\bp_t \Pi = (p_{t,h}\Pi_{h})_{h \in [H]}$ the stationary policy
such that, for all stages $h \in [H]$,
\begin{equation}
\label{eq:def-qPi}
p_{t,h}\Pi_{h} : s \in \cS \longmapsto p_{t,h}\Pi_{h}(\,\cdot\,|s) = \sum_{k \in [K]} p_{t,h}(k|s)\,\pi_{k,h}(\,\cdot\,|s) \in \cP(\cA)\,.
\end{equation}
Picking an action $a'$ according to $p_{t,h}\Pi_{h}(\,\cdot\,|s)$ amounts to performing
a two-stage randomization: first, drawing a policy index $k' \sim p_{t,h}(\,\cdot\,|s)$,
then drawing $a' \sim \pi_{k',h}(\,\cdot\,|s)$. This remark is important in the cases where
it is difficult or computationally complex to explicitly write the
$\pi_{k,h}(\,\cdot\,|s)$, but where it is easy to simulate them.

As indicated above, the set of all possible state-stage-dependent weights $\bq$ corresponds to $\cP\bigl([K]\bigr)^{[H] \times \cS}$.
We consider the class $\cC(\Pi)$ of all possible stationary policies defined according to~\eqref{eq:def-qPi}:
\[
\cC(\Pi) = \Bigl\{ \bq\Pi, \ \bq \in \cP\bigl([K]\bigr)^{[H] \times \cS} \Bigr\}\,,
\]
and aim to learn a good policy in this class.
To do so, the learning strategies pick weights $\bp_t \in \cP\bigl([K]\bigr)^{[H] \times \cS}$ over time and
output $\bpi_t = \bp_t \Pi$.
We will minimize the corresponding regret criterion:
\[
\forall T \geq 1, \qquad R^\Pi_T = \max_{\bq} \sum_{t=1}^T \Bigl( V_1^{\bq\Pi,\bcR_t}(s_1) - V_1^{\bp_t \Pi,\bcR_t}(s_1) \Bigr)\,.
\]

\begin{remark}
\label{rk:imitguar}
To the best of our understanding, the recent contributions
by \citet{CKA20} and \citet{Liu23} mentioned above
consider a more restrictive setting with a constant reward function and,
in addition, target a weaker notion of regret, corresponding to
\[
\max_{k \in [K]} V_1^{\bdelta_k\Pi,\bcR}(s_1) - \max_{t \in [T]} V_1^{\bp_t\Pi,\bcR}(s_1)\,,
\]
where each $\bdelta_k$ is a collection of state-stage-dependent weights that are all given by Dirac masses on expert~$k$; i.e.,
$V_1^{\bdelta_k\Pi,\bcR} = V_1^{\bpi_k,\bcR}$.
\end{remark}

Actually, the total regret $R_T$ defined in Section~\ref{sec:direct}
may be decomposed into some approximation
error, i.e., how good the policies in $\cC(\Pi)$ are in terms of values,
plus the regret with respect to $\cC(\Pi)$:
\[
R_T = \max_{\bpi} \sum_{t=1}^T \Bigl( V_1^{\bpi,\bcR_t}(s_1) - V_1^{\bpi_t,\bcR_t}(s_1) \Bigr) \\
= \underbrace{\max_{\bpi} \sum_{t=1}^T V_1^{\bpi,\bcR_t}(s_1) - \max_{\bq} \sum_{t=1}^T V_1^{\bq\Pi,\bcR_t}(s_1)}_{\mbox{\small approximation
error}} \ + \ R^\Pi_T\,.
\]
In this section, we aim to control $R^\Pi_T$ only and will assume that the approximation error is small
due to a proper choice of~$\Pi$. This situation is expected to arise frequently, as explained in
the following remark.

\begin{remark}
Denote by $\bpi^\star$ a stationary policy achieving the maximum in the
definition of $R_T$. Given that expert policies are combined through state-stage-dependent weights,
the approximation error defined above is null as soon as
\[
\forall h \in [H], \ \forall s \in \cS, \quad \exists q_h(\,\cdot\,|s) \in \cP\bigl([K]\bigr) \quad \mbox{s.t.} \quad
\pi^\star_h(\,\cdot\,|s) = \sum_{k \in [K]} q_h(k|s)\,\pi_{k,h}(\,\cdot\,|s)\,.
\]
In particular, it suffices that there exists $j^\star_{h,s} \in [K]$ such that $\pi^\star_h(\,\cdot\,|s) = \pi_{j^\star_{h,s},h}(\,\cdot\,|s)$.
Put differently, it suffices that at each stage $h \in [H]$ and for each state $s \in \cS$,
one of the expert policies (but not necessarily always the same) coincides with an optimal policy. This observation
motivates the use of expert policies in the cases where finitely many easy-to-identify distributions are candidates to
be optimal distributions for each given stage-state pair $(h,s)$.
\end{remark}

\paragraph{Summary.}
We provide in Box~B a summary of the settings and aims considered, here in Section~\ref{sec:aggreg}
and earlier in Section~\ref{sec:direct} (the left-hand side of Box~B corresponds to Box~A of
Section~\ref{sec:direct}).

\newcommand{\nboxtwocol}[3]{
\begin{tabular}{l|cl}
\multicolumn{3}{c}{ \ } \\[-.1cm]
\begin{minipage}[t]{0.465\textwidth}
{#1}
\end{minipage}
&
&
{#3}
\begin{minipage}[t]{0.48\textwidth}
{#2}
\end{minipage} \\[-0.25cm]
& & \\
\end{tabular}
}

\begin{figure}[t]
\begin{nbox}[title={Box~B: Policy optimization, possibly based on expert policies}]
\ \\[-.15cm]
\nboxtwocol{\underline{Direct tabular learning} (Section~\ref{sec:direct})}{\underline{Aggregation of expert policies} (Section~\ref{sec:aggreg})}{} \medskip

\hspace{2.75cm}\textbf{MDP parameters:} state space $\cS$, action space $\cA$, initial state $s_1 \in \cS$, \\
\phantom{\textbf{MDP parameters:}}\hspace{2.75cm} transition kernels $\bcT$ \medskip

\nboxtwocol{(\emph{No additional parameters})}{Set $\Pi$ of $K$ expert policies}{}
\ \\

\hspace{2.75cm} The environment picks a sequence $(\bcR_t)_{t \geq 1}$ of reward functions \\

\hspace{2.75cm}\textbf{For episodes} $t = 1, 2, \ldots$\textbf{:}
\begin{enumerate}[leftmargin=3.75cm,itemsep=1pt,topsep=3pt]
\item The initial state is set to $s_{t,1} = s_1$
\item \textbf{For stages} $h = 1, \ldots, H$\textbf{:}
\end{enumerate}
\ \\[-1.25cm]

\nboxtwocol{
\begin{enumerate}[label=(\alph*),topsep=0pt,itemsep=1pt,leftmargin=0.5cm]
\item The learner picks a policy $\pi_{t,h} : \cS \to \cP(\cA)$
\item and draws an action $a_{t,h} \sim \pi_{t,h}(\,\cdot\,|s_{t,h})$
\end{enumerate}
}{
\begin{enumerate}[label=(\alph*),topsep=0pt,itemsep=1pt,leftmargin=0.5cm]
\item The learner picks weights $p_{t,h} \in \cP\bigl([K]\bigr)^{\cS}$,
\item draws $k_{t,h} \sim p_{t,h}(\,\cdot\,|s_{t,h})$, the index
of the \\ expert policy,
\item and draws an action $a_{t,h} \sim \pi_{k_{t,h},h}(\,\cdot\,|s_t)$ \\
according to expert policy $k_{t,h}$
\end{enumerate}
}{}
\begin{enumerate}[leftmargin=2.75cm,itemsep=1pt,topsep=3pt]
\setcounter{enumi}{3}
\item The learner receives and observes a reward drawn independently \\ from $\cR_{t,h}(s_{t,h},a_{t,h})$,
with conditional expectation $r_{t,h}(s_{t,h},a_{t,h})$
\item If $h \leq H-1$, the next state $s_{t,h+1} \sim \cT_{h}(\,\cdot\,|s_{t,h},a_{t,h})$ is drawn
\end{enumerate}

\ \\

\hspace{5.75cm}\textbf{Goal:} Minimize the regret

\nboxtwocol{
$\displaystyle{R_T = \max_{\bpi} \sum_{t=1}^T \Bigl( V_1^{\bpi,\bcR_t}(s_1) - V_1^{\bpi_t,\bcR_t}(s_1) \Bigr)}$
}{
$\displaystyle{R^\Pi_T = \max_{\bq} \sum_{t=1}^T \Bigl( V_1^{\bq\Pi,\bcR_t}(s_1) - V_1^{\bp_t \Pi,\bcR_t}(s_1) \Bigr)}$
}{\phantom{T}\hspace{-.75cm}}
\end{nbox}
\end{figure}

\subsection{Equivalence between direct tabular learning and aggregation of expert policies}
\label{sec:eqv-imitation}

We now explain why any learning scheme minimizing the standard regret $R_T$
induces a learning scheme minimizing the regret $R^\Pi_T$ with respect to a finite set $\Pi$ of expert policies,
and vice versa.
In a nutshell, the equivalence stems from considering the indexes $k \in [K]$ of expert policies as
meta-actions, i.e., actions in a sequence of lifted MDPs.

As a consequence, for the sake of clarity and completeness, we will re-state
the counterpart of our main result, Theorem~\ref{th:main},
in the setting of policy orchestration: see Section~\ref{sec:advL-adv-bis}.
For now, we prove the claimed equivalence.

\paragraph{Direct tabular learning as aggregation of expert policies.}
We set $K = A$ and take as expert policies the Dirac masses on
the arms; more precisely, for each $a \in \cA$,
and for all $h \in [H]$ and $s \in \cS$, we set $\pi_{a,h}(\,\cdot\,|s) = \delta_a$, the Dirac mass at $a$.
This defines the expert policy $\bDelta_a$.
We consider
\[
\Delta = \{ \bDelta_a : a \in \cA \}
\qquad \mbox{and} \qquad
\cC(\Delta) = \bigl\{ \bp\Delta, \ \bp \in \cP(\cA)^{[H] \times \cS} \bigr\}\,;
\]
$\cC(\Delta)$ is the set of all stationary policies, stated in their direct tabular form.

\paragraph{From direct tabular learning to aggregation of expert policies.}
Conversely, we note that aggregation of expert policies in $\Pi$
amounts to performing direct tabular learning in the following sequence of (lifted) MDPs:
the action space is $\overline{\cA} = [K]$, the state space is $\overline{\cS} = \cS$,
the transition kernels $\overline{\bcT}$ and the reward functions $\overline{\bcR}_t$ are defined,
for all $t \geq 1$ and $h \in [H]$, by
\begin{align*}
& \overline{\cT}_h : (s,k) \in \cS \times [K] \longmapsto \sum_{a \in \cA} \pi_{k,h}(a|s)\,\cT_{h}(\,\cdot\,|s,a) \\
\mbox{and} \qquad &
\overline{\cR}_{t,h} : (s,k) \in \cS \times [K] \longmapsto \sum_{a \in \cA} \pi_{k,h}(a|s)\,\cR_{t,h}(s,a)\,.
\end{align*}
Direct tabular learning on the sequence of lifted MDPs defined above
provides policies $\overline{\bpi}_t$ which correspond to the convex
weights $\bp_t$ discussed above: for all $t \geq 1$, $h \in [H]$, and $s \in \cS$,
we use $p_{t,h}(\,\cdot\,|s) = \overline{\pi}_{t,h}(\,\cdot\,|s)$ to aggregate expert policies
in the original MDP.
Denoting by $\overline{R}_T$ the regret suffered with direct tabular learning in the lifted MDP,
we have: $R_T^\Pi = \overline{R}_T$.

\begin{remark}
In the final part of the proof of Theorem~\ref{th:main}, we critically used that
the transition kernels $\bcT$ do not depend on time.
The expression above for $\overline{\bcT}$ is indeed independent on time,
which would not be the case if the expert policies were evolving over time. This
explains why we restricted our attention to stationary expert policies.
\end{remark}

\subsection{Adversarial learning on advantage functions for aggregation of expert policies}
\label{sec:advL-adv-bis}

The counterpart for aggregation of expert policies of the strategy defined in Section~\ref{sec:advL-adv} is defined as follows,
given the equivalence stated above.

For each stage $h \in [H]$,
we fix a sequential strategy $\varphi_h = (\varphi_{t,h})_{t \geq 1}$ in the adversarial setting,
relying on reward vectors bounded by $M_h = H-h+1$ and of dimension $K$.

We run these strategies on the advantage functions of the lifted MDPs described above:
for all $t \geq 1$, $h \in [H]$, and $s \in \cS$,
\begin{equation}
\label{eq:def-olA-bis}
\overline{A}_h^{\bp_t,\overline{\bcR}_t}(s,\,\cdot\,)
= \Bigl( \overline{A}_h^{\bp_t,\overline{\bcR}_t}(s,k) \Bigr)_{k \in [K]}\,, \qquad \mbox{where} \qquad
\overline{A}_h^{\bp_t,\overline{\bcR}_t}(s,k) = \sum_{a \in \cA} \pi_{k,h}(a|s) \, A_h^{\bp_t\Pi,\bcR_t}(s,a)\,.
\end{equation}
More precisely, we run the strategies $(\varphi_h)_{h \in [H]}$
in the following stage-by-stage and state-by-state manner: for all $t \geq 1$,
\begin{equation}
\label{eq:ad-adv-def-aggr}
p_{t,h}(\,\cdot\, |s) = \varphi_{t,h} \Bigl( \bigl( \overline{A}_h^{\bp_\tau,\overline{\bcR}_\tau}(s,\,\cdot\,) \bigr)_{\tau \leq t-1} \Bigr)\,.
\end{equation}
We refer to this strategy as
$(\varphi_h)_{h \in [H]}$--\texttt{Adv2-Aggr}, for $(\varphi_h)_{h \in [H]}$--\underline{adv}ersarial learning on \underline{adv}antage functions
for \underline{aggr}egation of expert policies.

Theorem~\ref{th:main} immediately entails the following performance guarantee,
given the equivalence proved in Section~\ref{sec:eqv-imitation}.

\begin{corollary}
\label{cor:main}
In the setting of Section~\ref{sec:aggreg} where rewards lie in $[0,1]$,
if, for all $h \in [H]$, the sequential strategies $\varphi_h$ control the regret
in the adversarial setting (Definition~\ref{def:adv}) by $B_{T,K}$ for $K$--dimensional
reward vectors bounded by $H-h+1$,
then the $(\varphi_h)_{h \in [H]}$--\texttt{\emph{Adv2-Aggr}} strategy defined in~\eqref{eq:ad-adv-def-aggr}
over the set $\Pi$ of $K$ expert policies controls the regret with respect to $\cC(\Pi)$ as:
\[
\forall T \geq 1, \qquad
R^\Pi_T = \max_{\bq} \sum_{t=1}^T \Bigl( V_1^{\bq\Pi,\bcR_t}(s_1) - V_1^{\bp_t \Pi,\bcR_t}(s_1) \Bigr)
\leq H(H+1)\,B_{T,K}\,.
\]
\end{corollary}

\section{Empirical impacts as future research directions}
\label{sec:EstAdv}

\contrib{We review in greater detail how the literature
resorted or should resort to adversarial learning strategies in practice:
value functions are typically not observed and must be estimated.
These considerations call for future empirical research.}

\begin{figure}[t]
\center\fbox{
\begin{tabular}{ccc}
\emph{Strategy by \citet{Shani20}} & & \emph{Alternative formulations} \\
& & \\
\includegraphics[scale=0.94]{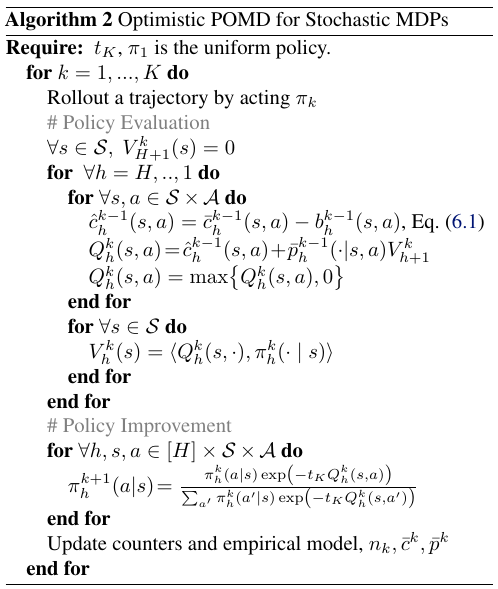} &
\begin{minipage}{0.02\textwidth}
\vspace{-3cm} \hspace{-.5cm}
$\boldsymbol{\longrightarrow}$
\end{minipage}
&
\begin{minipage}{0.385\textwidth}
\vspace{-4cm}

{\small
\textbf{for} all $(h,s) \in [H] \times \cS$ \textbf{do} \\
\[
\pi_{h}^{k+1}(\,\cdot\,|s) = \varphi_{k+1,h} \Bigl( \bigl( - Q_h^\tau(s,\,\cdot\,) \bigr)_{\tau \leq k} \Bigr)
\]
or, based on estimated advantage functions defined as in Assumption~\ref{ass:estA}:
\[
\pi_{h}^{k+1}(\,\cdot\,|s) = \varphi_{k+1,h} \Bigl( \bigl( - A_h^\tau(s,\,\cdot\,) \bigr)_{\tau \leq k} \Bigr)
\]
\textbf{end for}
}
\end{minipage}
\end{tabular}
}
\caption{\label{fig:Shani} The strategy considered and studied by \citet{Shani20},
as stated therein (\emph{left part}): our results focus on considering alternative formulations of the
policy-improvement step, based on other adversarial-learning strategies
than exponential weights, and possibly based on estimated advantage
functions rather than estimated $Q$--values (\emph{right part}).
\citet{Shani20} considers costs instead of rewards, hence the negative signs appearing
when feeding adversarial learning strategies $\varphi$ designed for
rewards.}
\end{figure}

This final section relaxes the assumption of an oracle providing value functions, as
stated at the beginning of Section~\ref{sec:advL-adv}, and first
recalls how the literature (see, e.g., \citealp{Politex19}, \citealp{Shani20}, \citealp{Cai20}, \citealp{He22}, \citealp{Zhao23})
typically performs the policy-improvement step for learning adversarial MDPs:
by resorting to the exponential-weight strategy of Section~\ref{sec:NPG}
based on estimated $Q$--values.
We provide a concrete example in Figure~\ref{fig:Shani}.

More precisely, the most popular approach is to build optimistic estimates
$\wh{Q}_h^t$ of the true $Q$--value functions $Q_h^{\bpi_t,\bcR_t}$, i.e., estimates
that upper bound the true values with high probability.
The main issue in doing so is that the transition
kernels $\bcT$ are unknown; whether the reward functions $\bcR_t$ are fully revealed (full-information feedback) or not (bandit feedback, where only actual rewards are observed) at the end of an episode may be handled
(see, among others, \citealp{Shani20}).
Also, these optimistic estimates $\wh{Q}_h^t$ may or may not rely on structural assumptions (e.g., \citealp{Cai20} and
\citealp{He22} assume some linear representation of the transition kernels)
and are specific to each article mentioned.

However, what is common to these articles, is the way the policy-improvement
step is performed based on these estimated $Q$--values;
this way is illustrated in Figure~\ref{fig:Shani}.
\citet{Politex19} even states that
``the choice of the Boltzmann policy is not arbitrary''
in this step (where ``Boltzmann policy'' is a synonym for
the exponential-weight strategy).

\emph{The point of the present article is exactly to question this common
practice and show that other choices are possible} (as detailed in Section~\ref{sec:EstAdv-1}),
\emph{with no impact on the theoretical guarantees proved in these articles}
(see Section~\ref{sec:EstAdv-2}). \emph{These observations call for
future empirical research} (see Section~\ref{sec:EstAdv-3}).

\subsection{Other choices for the policy-improvement step typically considered in the literature}
\label{sec:EstAdv-1}

As stated above and illustrated in Figure~\ref{fig:Shani},
the literature on adversarial MDPs (see, e.g., \citealp{Politex19}, \citealp{Shani20}, \citealp{Cai20}, \citealp{He22}, \citealp{Zhao23})
typically considers policy-improvement steps of the form, for some $\eta > 0$,
\begin{multline}
\label{eq:EWA-hatQ}
\forall t \geq 1, \ \
\forall h \in [H], \ \
\forall s \in \cS, \ \
\forall a \in \cA, \\
\pi_{t+1,h}(a|s) = \frac{\pi_{t,h}(a|s) \, \exp\bigl( \eta \, \hat{Q}_h^t(s,a) \bigr)}{\displaystyle{
\sum_{a' \in \cA} \pi_{t,h}(a'|s) \, \exp\bigl( \eta \, \hat{Q}_h^t(s,a') \bigr)}}
=
\frac{\displaystyle{\exp \!\left( \eta \sum_{\tau=1}^{t} \hat{Q}_h^\tau(s,a) \right)}}{
\displaystyle{\sum_{a' \in \cA} \exp \!\left( \eta \sum_{\tau=1}^{t} \hat{Q}_h^\tau(s,a') \right)}}\,.
\ \ \ \ ~
\end{multline}

We propose alternative formulations, based on advantage functions
obtained from the $Q$--values as described by Assumption~\ref{ass:estA}.
This assumption is satisfied as soon as value functions are also
obtained by the same convex combinations, which is the case
in all references mentioned above, but would not be the case for
other approaches, for instance, for $Q$--learning or similar methods
that would define
$\wh{V}^t_h(s)$ as $\displaystyle{\max_{a' \in \cA} \wh{Q}^t_h(s,a')}$.

Assumption~\ref{ass:estA} also imposes boundedness of the estimated value functions,
as this is key for the theoretical analysis of performance (and is coherent with
the fact that the true value functions are also bounded, by known bounds).

\begin{assumption}
\label{ass:estA}
The estimates $\wh{A}^t_h$ are defined based on estimates $\wh{Q}^t_h$ of $Q$--value functions,
using the policy $\bpi_t$ selected:
for all $s \in \cS$ and $a \in \cA$,
\[
\wh{V}^t_h(s) \eqdef \sum_{a' \in \cA} \pi_{t,h}(a'|s) \, \wh{Q}^t_h(s,a')
\qquad \mbox{and} \qquad
\wh{A}^t_h(s,a) \eqdef \wh{Q}^t_h(s,a) - \wh{V}^t_h(s)\,.
\]
In addition, the estimates $\wh{Q}^t_h$ and $\wh{A}^t_h$ are bounded, i.e.,
$0 \leq \wh{Q}^t_h \leq M_H$ and
$\bigl| \wh{A}^t_h \bigr| \leq M_H$ for some quantity $M_H$ (typically depending
on~$H$).
\end{assumption}

Formally, we propose to replace updates of the form~\eqref{eq:EWA-hatQ}
with the updates based on the strategy \texttt{Adv2} as stated in~\eqref{eq:ad-adv-def},
or its variant based on $Q$--values stated in~\eqref{eq:ad-adv-def-Q}:
\begin{align*}
& \forall t \geq 1, \ \
\forall h \in [H], \ \
\forall s \in \cS, \qquad
\pi_{t,h}(\,\cdot\, |s) = \varphi_{t,h} \Bigl( \bigl( \hat{A}_h^\tau(s,\,\cdot\,) \bigr)_{\tau \leq t-1} \Bigr) \\
\mbox{or} \qquad &
\forall t \geq 1, \ \
\forall h \in [H], \ \
\forall s \in \cS, \qquad
\pi_{t,h}(\,\cdot\, |s) = \varphi_{t,h} \Bigl( \bigl( \hat{Q}_h^\tau(s,\,\cdot\,) \bigr)_{\tau \leq t-1} \Bigr)\,.
\end{align*}
Empirically, updates based on estimated advantage functions should perform better.

\subsection{Preserved theoretical guarantees with these alternative choices}
\label{sec:EstAdv-2}

When the sequential strategies $\varphi_h$ control the regret
in the adversarial setting (Definition~\ref{def:adv}) by $B_{T,A}$ for $A$--dimensional
reward vectors, the same argument as in~\eqref{eq:csq-agreg-A}, together with
Assumption~\ref{ass:estA} for the equality to~$0$, shows that the strategies defined above satisfy:
for all $h \in [H]$ and $s \in \cS$,
\[
\max_{a \in \cA} \sum_{t=1}^T \wh{A}^t_h(s,a)
- \sum_{t=1}^T \overbrace{\sum_{a \in \cA} \pi_{t,h}(a|s) \, \wh{A}^t_h(s,a)}^{=\,0}
\leq 2 M_H\,B_{T,A}\,,
\]
thus, for all $h \in [H]$ and $s \in \cS$,
\begin{equation}
\label{eq:whRT}
\wh{R}_T \eqdef \max_{\bpi}
\sum_{h=1}^H \sum_{s \in \cS} \mu_h^{s_1,\bpi,\bcT}(s)
\sum_{a \in \cA} \pi_h(a|s) \sum_{t=1}^T \wh{A}^t_h(s,a) \leq 2 M_H\,B_{T,A}\,.
\end{equation}
Specific arguments (see detail below) then relate the quantity above to the target quantity,
stated as in~\eqref{eq:PDL-main}:
\begin{equation}
\label{eq:RT}
R_T = \max_{\bpi}
\sum_{t=1}^T \Bigl( V_1^{\bpi,\bcR_t}(s_1) - V_1^{\bpi_t,\bcR_t}(s_1) \Bigr)
= \max_{\bpi} \sum_{h=1}^H \sum_{s \in \cS} \mu_h^{s_1,\bpi,\bcT}(s) \sum_{a \in \cA} \pi_h(a|s) \sum_{t=1}^T A^{\bpi_t,\bcR_t}_{h}(s,a)\,.
\end{equation}

\paragraph{Typical examples.}
For instance, in the theoretical analysis by \citet[Section~6]{Shani20},
the total regret $R_T$ is decomposed as a sum of three terms,
where term~(ii) therein is exactly~\eqref{eq:whRT}
but terms~(i) and~(iii) are bounded in some specific way.

The same may be mentioned for the other references,
where the total regret $R_T$ is also decomposed into three terms,
with $\wh{R}_T$ being one of the three terms:
in \citet{Cai20}, term~(i);
in \citet{He22}, term~$I_1$;
in \cite{Zhao23}, the ``OMD regret term''.

Again, the adversarial-learning strategy $(\varphi_h)_{h \in [H]}$
considered in all these references is the exponential potential with a constant learning rate
(see Section~\ref{sec:NPG}), possibly seen as an instance of online mirror descent,
and bounds of typical order $\sqrt{T \ln A}$ are achieved on the term corresponding to $\wh{R}_T$.
Thus, considering other adversarial-learning strategies,
in the forms described in Section~\ref{sec:EstAdv-1}, would not hurt
the final regret bounds achieved on $R_T$, as the errors stemming from the control
of $\wh{R}_T$ are not the main contributors to the final regret bounds
achieved in these articles.

\paragraph{An exception.}
In general, $\wh{R}_T$ is not equal to a sum of differences of value functions.
An exception is to be found in \citet{Tiapkin25}: they obtain the estimates~$\wh{Q}^t_h$
as the \emph{exact} $Q$--value functions (obtained by dynamic programming) corresponding to the policies $\bpi_t$, to some reward functions $\bcR'_t$
(based on the actual reward function $\bcR_t$ revealed at the end of the episode plus some bonus function),
and to some estimated transition kernels $\wh{\bcT}_t$ (that are constant over subintervals of episodes
and are only updated from time to time).
\citet{Tiapkin25} decompose the total regret in four terms, where term~(B) corresponds to $\wh{R}_T$.

Note that \citet{Tiapkin25} refers to the present work and is therefore able to present
the analysis in a more modular way than many of the references mentioned above
(some of them re-deriving regret guarantees in terms of $Q$--value functions by mimicking
proofs of adversarial regret bounds for exponential weights, as reviewed in Section~\ref{sec:lit-review}).

\subsection{Future research: evaluation of the empirical impacts of these alternative choices}
\label{sec:EstAdv-3}

We explained in Section~\ref{sec:EstAdv-2} that the modifications proposed
in Section~\ref{sec:EstAdv-1} for the policy-improvement step
preserve the theoretical guarantees
(essentially because this policy-improvement step is not at all the main blocking point in the analysis).

Yet, a different formulation of this policy-improvement step
may dramatically affect the practical performance obtained by these strategies.
Many adversarial-learning strategies exist, and exponential-weight strategies
are often not the best-performing ones.
The \texttt{R} package \texttt{Opera} by \citet{CRAN}
implements several adversarial-learning strategies, some of which (for instance, \texttt{ML-Prod} and \texttt{ML-Poly},
introduced by \citealp{Gai14})
often achieving superior empirical performance compared to exponential weights
(see the empirical studies in \citealp{Gai15}).

Therefore, in our opinion, an interesting avenue of empirical research would
be the following. Consider the experimental designs provided by, or create experimental designs
corresponding to, the settings of \citet{Politex19}, \citet{Shani20}, \citet{Cai20}, \citet{He22}, \citet{Zhao23}.
Compare the performance of the strategies introduced therein (with their original policy-improvement steps,
based on exponential weights) to alternative strategies differing only in their policy-improvement steps (based on the
adversarial-learning strategies implemented in the \texttt{Opera} package, used either
with estimated $Q$--values or with estimated advantage functions).
Also, for exponential weights, as mentioned after Corollary~\ref{cor:NPG}, it would be interesting
to see the effect of the learning rate $\eta$ on the empirical performance.
We leave these studies for future research.

\bibliography{Aggreg-RL--TMLR--bib}
\bibliographystyle{tmlr}

\appendix

\paragraph{Appendix.} The appendix provides proofs omitted from the main body of the article.

\section{Proof of Theorem~\ref{th:NPG} (analysis of NPG with softmax parametrization)}
\label{app:NPG}

For the convenience of the reader, we restate the result to be proved.

\thNPG*

As indicated in Section~\ref{sec:NPG}, the proof below is based on
the analysis of the natural policy gradient [NPG]
with softmax parametrization proposed by \citet[Section~5.3]{AKLM21}
in the discounted setting with reward functions constant over time.
See Remark~\ref{rk:proof-diff} for an explanation of why the proof
in the discounted setting is significantly simpler than the proof in the episodic setting.

We extend the proof of \citet[Section~5.3]{AKLM21} to the episodic setting
and to (obliviously) adversarial sequences of reward functions.
We also claim a more transparent proof scheme, consisting
of an ad hoc adversarial bound (Lemma~\ref{lm:NPG}) which is then
applied to policy learning along the lines of the proof of Theorem~\ref{th:main}.

More precisely, the first piece of the proof of Theorem~\ref{th:NPG} is to
replace the uniform regret bounds considered in Definition~\ref{def:adv}
with some ad hoc, data-based, bound
(of the same flavor as the bounds by \citealp[Section~2]{FlipFlop14} in terms of so-called
mixability gaps). Indeed, the uniform regret bound
that could be proved (see, e.g., \citealp[Theorem~2.2]{CBL06})
for the adversarial strategy of Lemma~\ref{lm:NPG}
is $B_{T,K} = \ln K/\eta + \eta T/ 8$, which is not sublinear.

\begin{lemma}
\label{lm:NPG}
The strategy~\eqref{eq:def-pot-based} based on a constant exponential potential $\Phi_t \equiv \Phi : x \mapsto \exp( \eta x)$,
i.e., picking weights
\[
\forall t \geq 1, \qquad
w_{t,k} = \frac{v_{t,k}}{\displaystyle{\sum_{j \in [K]} v_{t,j}}} \,, \qquad \mbox{where} \qquad
v_{t,k} = \exp \!\left( \eta \sum_{\tau=1}^{t-1} g_{\tau,k} \right),
\]
with the convention that $v_{1,k} = 1$ and $w_{1,k} = 1/K$,
satisfies the following bound: against all opponents sequentially picking
reward vectors in $\R^K$,
\[
\forall T \geq 1, \qquad
\max_{k \in [K]} \sum_{t=1}^T g_{t,k}
\leq \frac{\ln K}{\eta} + \sum_{t=1}^T \sum_{j \in [K]} w_{t+1,j} \, g_{t,j}\,.
\]
\end{lemma}

This lemma is proved at the end of this section and we now
apply it to prove Theorem~\ref{th:NPG}.

\begin{proof}[Proof of Theorem~\ref{th:NPG}]
We adapt the proof of Theorem~\ref{th:main} by replacing~\eqref{eq:csq-agreg-A} by the ad hoc bound
stemming from Lemma~\ref{lm:NPG}; we obtain:
\begin{equation}
\label{eq:NPG-lemma}
\forall h \in [H], \ \
\forall s \in \cS, \qquad
\max_{a \in \cA} \sum_{t=1}^T A^{\bpi_t,\bcR_t}_{h}(s,a)
\leq \frac{\ln A}{\eta} + \sum_{t=1}^T \sum_{a \in \cA} \pi_{t+1,h}(a|s) \, A^{\bpi_t,\bcR_t}_{h}(s,a)
\,.
\end{equation}
We fix a comparator policy $\bpi$.
The combination of the obtained inequality~\eqref{eq:NPG-lemma} with
the application~\eqref{eq:PDL-main} of the performance difference lemma yields
\begin{align}
\nonumber
\sum_{t=1}^T \Bigl( V_1^{\bpi,\bcR_t}(s_1) - V_1^{\bpi_t,\bcR_t}(s_1) \Bigr)
& \leq \sum_{h=1}^H \sum_{s \in \cS} \mu_h^{s_1,\bpi,\bcT}(s) \,\, \max_{a \in \cA} \sum_{t=1}^T A^{\bpi_t,\bcR_t}_{h}(s,a) \\
\label{eq:PDL-NPG-debut}
& \leq \frac{H \ln A}{\eta} + \sum_{t=1}^T \sum_{h=1}^H \, \underbrace{\sum_{s \in \cS}
\mu_h^{s_1,\bpi,\bcT}(s) \sum_{a \in \cA} \pi_{t+1,h}(a|s) \, A^{\bpi_t,\bcR_t}_{h}(s,a)}_{\mbox{\scriptsize to be bounded}}\,.
\end{align}
We fix $t \in [T]$ and $h \in [H]$ and
define a new one-shot policy $\tilde\bpi_{t+1}^h = \bigl( \tilde\pi_{t+1,h'}^h \bigr)_{h' \in [H]}$ as follows:
\begin{numcases}{\tilde\pi_{t+1,h'}^h =}
\nonumber \pi_{h'} & if $h' \leq h-1$, \\
\nonumber \pi_{t+1,h'} & if $h' \geq h$.
\end{numcases}
As $\bpi$ and $\tilde\bpi_{t+1}^h$ coincide in the first $h-1$ stages, we have
$\mu_h^{s_1,\bpi,\bcT}(s) = \mu_h^{s_1,\tilde\bpi_{t+1}^h,\bcT}(s)$.
In addition, the definition of $\tilde\bpi_{t+1}^h$,
the definition of the strategy, Lemma~\ref{lm:monot}, and the definition of advantage functions
entail that for all $s \in \cS$ and all $h' \geq h$,
\[
\sum_{a \in \cA} \tilde{\pi}_{t+1,h'}^h(a|s) \, A^{\bpi_t,\bcR_t}_{h'}(s,a) =
\sum_{a \in \cA} \pi_{t+1,h'}(a|s) \, A^{\bpi_t,\bcR_t}_{h'}(s,a) \geq
\sum_{a \in \cA} \pi_{t,h'}(a|s) \, A^{\bpi_t,\bcR_t}_{h'}(s,a) =
0\,.
\]
Therefore, the sum marked as ``to be bounded'' in~\eqref{eq:PDL-NPG-debut} can be controlled as
\begin{align}
\nonumber
\sum_{s \in \cS}
\mu_h^{s_1,\bpi,\bcT}(s) \sum_{a \in \cA} \pi_{t+1,h}(a|s) \, A^{\bpi_t,\bcR_t}_{h}(s,a)
& = \sum_{s \in \cS}
\mu_h^{s_1,\tilde\bpi_{t+1}^h,\bcT}(s) \sum_{a \in \cA} \tilde{\pi}_{t+1,h}^h(a|s) \, A^{\bpi_t,\bcR_t}_{h}(s,a) \\
\nonumber
& \leq \sum_{h' = h}^H \sum_{s \in \cS}
\mu_{h'}^{s_1,\tilde\bpi_{t+1}^h,\bcT}(s) \sum_{a \in \cA} \tilde{\pi}_{t+1,h'}^h(a|s) \, A^{\bpi_t,\bcR_t}_{h'}(s,a) \\
\label{eq:PDL-NPG-interm}
& = \sum_{s \in \cS} \mu_{h}^{s_1,\bpi,\bcT}(s) \Bigl( V_h^{\tilde\bpi_{t+1}^h,\bcR_t}(s) - V_h^{\bpi_t,\bcR_t}(s) \Bigr)\,,
\end{align}
where the final equality~\eqref{eq:PDL-NPG-interm} follows from the
equality of distributions $\smash{\mu_h^{s_1,\bpi,\bcT} = \mu_h^{s_1,\tilde\bpi_{t+1}^h,\bcT}}$
at stage $h$ (which holds because
$\bpi$ and $\tilde\bpi_{t+1}^h$ coincide in the first $h-1$ stages)
together with an application of
the performance difference lemma (Lemma~\ref{lm:PDL}).

As $\tilde\bpi_{t+1}^h$ and $\bpi_{t+1}$ coincide in the last $h$ stages, we have
$V^{\tilde\bpi_{t+1}^h,\bcR_t}_{h}(s) = V^{\bpi_{t+1},\bcR_t}_{h}(s)$ for all $s \in \cS$.
This observation, combined with~\eqref{eq:PDL-NPG-interm}, entails
\[
\sum_{s \in \cS}
\mu_h^{s_1,\bpi,\bcT}(s) \sum_{a \in \cA} \pi_{t+1,h}(a|s) \, A^{\bpi_t,\bcR_t}_{h}(s,a)
\leq \sum_{s \in \cS} \mu_{h}^{s_1,\bpi,\bcT}(s) \Bigl( V^{\bpi_{t+1},\bcR_t}_{h}(s) - V_h^{\bpi_t,\bcR_t}(s) \Bigr)\,,
\]
and we thus get, after substitution into~\eqref{eq:PDL-NPG-debut},
\begin{equation}
\label{eq:final-bd-NPG}
\sum_{t=1}^T \Bigl( V_1^{\bpi,\bcR_t}(s_1) - V_1^{\bpi_t,\bcR_t}(s_1) \Bigr)
\leq \frac{H \ln A}{\eta} + \sum_{h=1}^H
\sum_{s \in \cS} \mu_{h}^{s_1,\bpi,\bcT}(s) \sum_{t=1}^T
\Bigl( V^{\bpi_{t+1},\bcR_t}_{h}(s) - V_h^{\bpi_t,\bcR_t}(s) \Bigr) \,.
\end{equation}
We obtain telescoping sums on regimes of payoffs. More precisely,
with the notation~\eqref{eq:def-R},
\[
\forall k \in \{2,\ldots,R+1\}, \qquad
\sum_{t=\tau_{k-1}}^{\tau_k - 1} \Bigl( V^{\bpi_{t+1},\bcR_t}_{h}(s) - V_h^{\bpi_t,\bcR_t}(s) \Bigr)
= V^{\bpi_{\tau_k},\bcR_{\tau_{k-1}}}_{h}(s) - V^{\bpi_{\tau_{k-1}},\bcR_{\tau_{k-1}}}_{h}(s)
\leq H-h+1\,,
\]
where the upper bound follows from the boundedness of rewards in~$[0,1]$.
Together with~\eqref{eq:final-bd-NPG}, we finally obtain
\[
\sum_{t=1}^T \Bigl( V_1^{\bpi,\bcR_t}(s_1) - V_1^{\bpi_t,\bcR_t}(s_1) \Bigr)
\leq \frac{H \ln A}{\eta} + \sum_{h=1}^H
\sum_{s \in \cS} \mu_{h}^{s_1,\bpi,\bcT}(s) \sum_{k=2}^{R+1} (H-h+1)
= \frac{H \ln A}{\eta} + \frac{R H (H+1)}{2}\,,
\]
which leads to the claimed regret upper bound after taking the maximum
over all policies $\bpi$.
\end{proof}

\begin{remark}
\label{rk:proof-diff}
The arguments between~\eqref{eq:PDL-NPG-debut} and~\eqref{eq:final-bd-NPG}
may be bypassed in the discounted setting with discount factor~$\gamma$;
see \citet[Section~5.3]{AKLM21}. More precisely,
(with obvious notation, for value functions defined in the standard way for discounted rewards,
and for a constant reward function), for each $s \in \cS$,
\begin{multline*}
\max_{a \in \cA} \sum_{t=1}^T A^{\pi_t}(s,a)
\leq \frac{\ln A}{\eta} + \sum_{t=1}^T \overbrace{\sum_{a \in \cA} \pi_{t+1}(a|s) \, A^{\pi_t}(s,a)}^{\geq 0} \\
\leq \frac{\ln A}{\eta} + \sum_{t=1}^T \underbrace{\frac{1}{1-\gamma}
\sum_{s' \in \cS} \mu^{s,\pi_{t+1}}(s') \sum_{a \in \cA} \pi_{t+1}(a|s') \, A^{\pi_t}(s',a)}_{= V^{\pi_{t+1}}(s) - V^{\pi_{t}}(s)}
= \frac{\ln A}{\eta} + \underbrace{V^{\pi_{T+1}}(s) - V^{\pi_1}(s)}_{\leq 1/(1-\gamma)} \,,
\end{multline*}
where the first inequality is by Lemma~\ref{lm:NPG},
where the non-negativity is guaranteed by monotonicity of weights (see Lemma~\ref{lm:monot}),
where the second inequality comes from the fact
that distributions induced by a starting state~$s$, a given policy, and a given transition function
put a probability mass at least $1-\gamma$ on $s$, no matter the policy and transition function
(this is the property extremely specific to the discounted setting),
where the equality to $V^{\pi_{t+1}}(s) - V^{\pi_{t}}(s)$ is by
the performance difference lemma,
and where the final equality is by telescoping.
The inequality obtained above is the key;
the rest of the proof merely consists of yet another (now standard)
application of the performance difference lemma:
\[
\sum_{t=1}^T \Bigl( V^{\pi}(s_1) - V^{\pi_t}(s_1) \Bigr)
= \sum_{t=1}^T \frac{1}{1-\gamma} \sum_{s \in \cS} \mu^{s_1,\pi}(s) \sum_{a \in \cA} \pi(a|s) \, A^{\pi_t}(s,a)
\leq \frac{1}{1-\gamma} \sum_{s \in \cS} \mu^{s_1,\pi}(s)
\underbrace{\max_{a \in \cA} \sum_{t=1}^T A^{\pi_t}(s,a)}_{\leq (\ln A)/\eta + 1/(1-\gamma)}\,,
\]
which is the bound claimed by \citet[Section~5.3]{AKLM21}.
\end{remark}

We conclude this section with a proof of Lemma~\ref{lm:NPG}.

\begin{proof}[Proof of Lemma~\ref{lm:NPG}]
First, a bound ``{\`a} la Pisier'' yields that for all sequences of payoffs $g_{t,j}$, possibly signed
and unbounded:
\begin{align*}
\max_{k \in [K]} \sum_{t=1}^T g_{t,k}
& = \frac{1}{\eta} \ln \Biggl( \max_{j \in [K]} \exp \biggl( \eta \sum_{t=1}^T g_{t,j} \biggl) \Biggr) \\
& \leq \frac{1}{\eta} \ln \!\left( \sum_{j \in [K]} \exp \! \left( \eta \sum_{t=1}^T g_{t,j} \right) \right)
= \frac{\ln K}{\eta} + \frac{1}{\eta} \sum_{t=1}^T \ln \!\left( \sum_{j \in [K]} w_{t,j} \, \exp(\eta g_{t,j}) \right),
\end{align*}
where the equality follows by telescoping: indeed, by definition of the weights,
\[
\sum_{j \in [K]} \underbrace{\exp \! \left( \eta \sum_{t=1}^T g_{t,j} \right)}_{= v_{T+1,j}}
= K \prod_{t=1}^T \frac{\displaystyle{\sum_{j \in [K]} v_{t+1,j}}}{\displaystyle{\sum_{j \in [K]} v_{t,j}}}
= K \prod_{t=1}^T \frac{\displaystyle{\sum_{j \in [K]} v_{t,j}} \, \exp(\eta g_{t,j})}{\displaystyle{\sum_{j \in [K]} v_{t,j}}}
= K \prod_{t=1}^T w_{t,j} \, \exp(\eta g_{t,j})\,.
\]
Second, by the application of Jensen's inequality to the convex function $x \mapsto x \ln x$,
\[
\left( \sum_{j \in [K]} w_{t,j} \, \exp(\eta g_{t,j}) \right) \ln \!\left( \sum_{j \in [K]} w_{t,j} \, \exp(\eta g_{t,j}) \right)
\leq \sum_{j \in [K]} w_{t,j} \, \exp(\eta g_{t,j}) \ln \bigl( \exp(\eta g_{t,j}) \bigr)\,,
\]
that is, after rearranging and given the definition of the weights $w_{j,t+1}$,
\[
\ln \!\left( \sum_{j \in [K]} w_{t,j} \, \exp(\eta g_{t,j}) \right)
\leq \eta \sum_{j \in [K]} w_{t+1,j} \, g_{t,j} \,.
\]
The claimed bound follows from combining the two inequalities obtained.
\end{proof}

\section{Proof of the performance difference lemma}
\label{sec:PDL}

One of the first references stating the performance difference lemma (in the discounted setting)
is \citet{KL02}.
Statements (possibly of generalizations) of this lemma
for $H$--episodic MDPs are ubiquitous in the literature (see, e.g.,
\citealp[Lemma~3.2]{Cai20} for a simple statement,
and \citealp[Lemma~1]{Shani20} for an extension to approximated advantage functions).
We state yet another, straightforward, generalization, in terms
of advantage and value functions starting at a given stage~$h$;
this generalization is useful in the
proof of Theorem~\ref{th:NPG} in Appendix~\ref{app:NPG}.

\lmPDL*

\begin{proof}
We denote by $\P^{s_1,\bpi,\bcT}$ the probability distribution underlying
the $H$--episodic MDP $(s_1,a_1,\ldots,s_H,a_H)$
starting at $s_1$, drawing actions according to $\bpi$, and subject to the
transition kernels~$\bcT$. In particular, by definition, for any function
$f : \cS \times \cA \to \R$ and all $h' \in [H]$,
\[
\sum_{s \in \cS} \mu_{h'}^{s_1,\bpi,\bcT}(s) \sum_{a \in \cA} \pi_{h'}(a|s)\,f(s,a)
= \E^{s_1,\bpi,\bcT} \bigl[ f(s_{h'},a_{h'}) \bigr]\,.
\]
Letting successively $f$ be $ A^{\bpi',\bcR}_{h'}$ for $h \leq h' \leq H$ and
using the definition $A^{\bpi',\bcR}_{h'} = Q^{\bpi',\bcR}_{h'} - V^{\bpi',\bcR}_{h'}$,
\begin{equation}
\label{eq:lmPDF-telesc}
\sum_{h'=h}^H \sum_{s \in \cS} \mu_{h'}^{s_1,\bpi,\bcT}(s) \sum_{a \in \cA} \pi_{h'}(a|s) \, A^{\bpi',\bcR}_{h'}(s,a)
= \E^{s_1,\bpi,\bcT} \! \left[ \sum_{h'=h}^H \bigl( Q^{\bpi',\bcR}_{h'}(s_{h'},a_{h'}) - V^{\bpi',\bcR}_{h'}(s_{h'}) \bigr) \right] .
\end{equation}
Now, by definition of the $Q$--values,
recalling that $\br$ denotes the mean-payoff functions associated with $\bcR$, we have, for $h' \leq H-1$,
\begin{equation}
\label{eq:lmPDF-sa}
\forall (s,a) \in \cS \times \cA, \qquad
Q^{\bpi',\bcR}_{h'}(s,a) = r_{h'}(s,a) + \sum_{s' \in \cS} \cT_{h'}(s' \,|\, s,a) \, V^{\bpi',\bcR}_{h'+1}(s')\,.
\end{equation}
By definition of the MDP, for any function $g : \cS \to \R$,
\[
\E^{s_1,\bpi,\bcT} \! \left[ \sum_{s' \in \cS} \cT_{h'}(s' \,|\, s_{h'},a_{h'}) \, g(s') \right]
= \E^{s_1,\bpi,\bcT} \bigl[ g(s_{h'+1}) \bigr]\,.
\]
Thus, letting $s = s_{h'}$ and $a = a_{h'}$ in~\eqref{eq:lmPDF-sa} and taking expectations yields
\[
\E^{s_1,\bpi,\bcT} \bigl[ Q^{\bpi',\bcR}_{h'}(s_{h'},a_{h'}) \bigr] =
\E^{s_1,\bpi,\bcT} \bigl[ r_{h'}(s_{h'},a_{h'}) \bigr] +
\E^{s_1,\bpi,\bcT} \bigl[ V^{\bpi',\bcR}_{h'+1}(s_{h'+1}) \bigr] \, .
\]
For $h' = H$, we have $Q^{\bpi',\bcR}_{H}(s,a) = r_H(s,a)$.
As a consequence of the equalities above, a telescoping sum appears in the right-hand side of~\eqref{eq:lmPDF-telesc}:
\begin{align*}
& \E^{s_1,\bpi,\bcT} \! \left[ \sum_{h'=h}^H \bigl( Q^{\bpi',\bcR}_{h'}(s_{h'},a_{h'}) - V^{\bpi',\bcR}_{h'}(s_{h'}) \bigr) \right] \\
= \ & \E^{s_1,\bpi,\bcT} \! \left[ r_{h'}(s_H,a_H) + \sum_{h'=h}^{H-1} \bigl( r_{h'}(s_{h'},a_{h'}) + V^{\bpi',\bcR}_{h'+1}(s_{h'+1}) \bigr)
- \sum_{h'=h}^{H} V^{\bpi',\bcR}_{h'}(s_{h'}) \right] \\
= \ & \E^{s_1,\bpi,\bcT} \! \left[ \sum_{h'=h}^H r_{h'}(s_{h'},a_{h'}) \right]
- \E^{s_1,\bpi,\bcT} \bigl[ V^{\bpi',\bcR}_{1}(s_h) \bigr] \, .
\end{align*}
Finally, the tower rule shows that
\[
\E^{s_1,\bpi,\bcT} \! \left[ \sum_{h'=h}^H r_{h'}(s_{h'},a_{h'}) \right]
= \E^{s_1,\bpi,\bcT} \bigl[ V^{\bpi,\bcR}_{1}(s_h) \bigr] \,.
\]
The proof is concluded by collecting all the bounds.
\end{proof}

\end{document}